\documentclass[letterpaper,twocolumn,10pt]{article}
\usepackage{usenix2019_v3}
\usepackage{graphicx}
\usepackage{amsmath,amssymb}
\usepackage{booktabs,multirow}
\usepackage{tikz}
\usepackage{xspace}
\usepackage{wrapfig}
\usepackage{caption}
\usepackage{csquotes}
\usepackage[framemethod=tikz]{mdframed}
\PassOptionsToPackage{hyphens}{url}

\usepackage{titling}

\begin{document}

\newcommand\ds{\raisebox{-1pt}{\tikz \draw [line cap=round, line width=0.25ex, dash pattern=on 0pt off 2pt] (0,0) circle [radius=0.75ex];}}
\newcommand\ssn{\ds{}\ds{}\ds{}-\ds{}\ds{}-\ds{}\ds{}\ds{}\ds{}}
\newtheorem{pd}{Problem Definition}
\newtheorem{definition}{Definition}
\newtheorem{thm}{Theorem}

\newif\ifsubmit
\submittrue

\newif\ifalt
\alttrue

\newif\ifintrothree
\introthreetrue

\newcommand{\npc}[1]{{\color{blue}[npc: #1]}}
\newcommand{\dawn}[1]{{\color{red}[Dawn: #1]}}
\ifsubmit
\newcommand{\ulfar}[1]{}
\newcommand{\chang}[1]{}
\newcommand{\om}[1]{}
\newcommand{\richard}[1]{}
\newcommand{\todo}[1]{}
\newcommand{\eat}[1]{}
\newcommand{\fixlang}{}
\newcommand{\fixsq}{}
\else
\newcommand{\ulfar}[1]{{\color{blue}[Ulfar: #1]}}
\newcommand{\chang}[1]{{\color{blue}[Chang: #1]}}
\newcommand{\npc}[1]{{\color{blue}[npc: #1]}}
\newcommand{\om}[1]{{\color{blue}[Om: #1]}}
\newcommand{\richard}[1]{{\color{purple}[Richard: #1]}}
\newcommand{\todo}[1]{{\color{red}[TODO: #1]}}
\newcommand{\eat}[1]{}
\newcommand{\fixlang}{{\marginpar{TODO FIX LANG}}}
\newcommand{\fixsq}{{\marginpar{TODO SQUEEZ}}}
\newcommand{\dawn}[1]{{\color{red}[Dawn: #1]}}
\fi

\newcommand{\secret}{{format}\xspace}
\newcommand{\sent}{\mathrm{Px}}
\newcommand{\randomness}{{randomness}\xspace}
\newcommand{\entropy}{{log-perplexity}\xspace}
\newcommand{\entropyshort}{{SE}\xspace}
\newcommand{\he}{{high-entropy}\xspace}
\newcommand{\google}[1]{{\color{brown}PlaceHolder}}
\renewcommand{\paragraph}[1]{{\vspace{5pt}\noindent\bf #1}}

\newcommand{\secretX}{{\mathfrak{X}}}%

\date{}
\setlength{\droptitle}{-5em}
\title{\Large \bf The Secret Sharer:
Evaluating and Testing \\
Unintended Memorization 
in Neural Networks}

\author{
{\rm Nicholas Carlini\textsuperscript{1,}\textsuperscript{2}}\\
\and
{\rm Chang Liu\textsuperscript{2}}
\and
{\rm {\'U}lfar Erlingsson\textsuperscript{1}}
\and
{\rm Jernej Kos\textsuperscript{3}}
\and
{\rm Dawn Song\textsuperscript{2}}
\and
\textsuperscript{1}\emph{Google Brain} \,\,
\textsuperscript{2}\emph{University of California, Berkeley} \,\,
\textsuperscript{3}\emph{National University of Singapore}
} %
\date{\vspace{-2em}}

\maketitle
\thispagestyle{plain}
\pagestyle{plain}

\begin{abstract}
\ifintrothree
This paper describes a testing methodology
for quantitatively assessing
the risk that
rare or unique training-data sequences are
\emph{unintentionally memorized}
by generative sequence models---a
common type of machine-learning model.
Because such models are sometimes trained on sensitive data
(e.g., the text of users' private messages),
this methodology
can benefit privacy by
allowing deep-learning practitioners
to select means of training that minimize such memorization.

In experiments, we show that unintended memorization
is a persistent, hard-to-avoid issue
that can have serious consequences.
Specifically, 
for models 
trained without consideration of memorization,
we describe new, efficient procedures 
that can extract
unique, secret sequences,
such as credit card numbers.
We show that our testing strategy is a practical and easy-to-use
first line of defense,
e.g., by describing its application to 
quantitatively limit data exposure in
Google's Smart Compose, a commercial text-completion neural network
trained on millions of users' email messages.

\else
Machine learning models based on neural networks
are being rapidly adopted for many purposes.
What details those models may have unintentionally memorized,
and may reveal,
can be of significant concern,
at least when the models are public
and trained on data that
includes
sensitive, private information.

In the context of generative sequence models,
this paper introduces
a quantitative metric by which
practitioners can measure their models' propensity 
for exposing details about private training data.
Our metric, called \emph{exposure}, 
can be applied during training,
as part of a structured testing strategy,
to empirically measure
a model's potential for unintended memorization
of unique or rare sequences in the training data.

In evaluating our metric,
we find unintended memorization
to be both commonplace and hard to prevent.
In particular,
such memorization is not due to overfitting by training too long:
it occurs early during training,
and persists across different types
of models, hyperparameters, and training strategies.
Furthermore,
the use of simple, intuitive 
approaches such as regularization
(e.g., dropout)
is insufficient to prevent
unintended memorization.
Only by using
differentially-private training techniques
can we eliminate the issue completely,
albeit at some loss in utility.

Our exposure-based testing strategy is practical, 
as we demonstrate in experiments,
and by describing its use
in removing privacy risks 
for a deployed, commercial model
that is trained on millions of users' email messages
and used to predict sentence completion
during (other) users' email composition.

Finally,
to
motive the use of our testing strategy,
we describe an algorithm
that, given access to a trained model,
can efficiently extract unique, secret sequences
for any known sequence prefix,
guided by our exposure metric.
We demonstrate our algorithm's
effectiveness in experiments,
e.g., by extracting credit card numbers
from a language model
trained on the Enron email data.
Such empirical extraction
can usefully 
convince practitioners
that unintended memorization is not just of academic interest,
but an issue of serious, practical concern.
\fi
\end{abstract}

\pagestyle{empty}
\thispagestyle{empty}

\ifintrothree
\section{Introduction}
When a secret is shared,
it can be very difficult to prevent its further disclosure---as
artfully explored in Joseph Conrad's
\textit{The Secret Sharer}~\cite{secretsharer}.
This difficulty also arises in 
machine-learning models based on neural networks,
which are being rapidly adopted for many purposes.
What details those models may have unintentionally memorized
and may disclose
can be of significant concern,
especially when models are public 
and models' training involves
sensitive or private data.

Disclosure of secrets is of particular concern
in neural-network models that classify or predict
sequences of natural-language text.
First, 
such text
will often contain
sensitive or private sequences,
accidentally,
even if the text is supposedly public.
Second, 
such models are designed to learn
text patterns such as grammar, turns of phrase,
and spelling, %
which 
comprise a vanishing fraction
of the exponential space of all possible sequences.
Therefore, 
even if sensitive or private training-data text is very rare,
one should assume that well-trained models  have paid attention to its precise details.

Concretely, 
disclosure of secrets may arise naturally in
generative text models like those
used for text auto-completion and predictive keyboards,
if trained on possibly-sensitive data.
The users of such models may
discover---either by accident or on purpose---that
entering certain text prefixes causes
the models to output 
surprisingly-revealing text completions \cite{xkcd}.
For example,
users may find that the input 
``my social-security number is\ldots{}''
gets auto-completed to an obvious secret
(such as a valid-looking SSN not their own),
or find that other inputs 
are auto-completed to text with oddly-specific details.
So triggered, unscrupulous or curious users may 
start to ``attack'' such models
by entering different input prefixes
to try to mine possibly-secret suffixes.
Therefore, 
for generative text models,
assessing and reducing the chances 
that secrets may be disclosed
in this manner is
a key practical concern.

To enable
practitioners to measure their models' propensity 
for disclosing details about private training data,
this paper introduces 
a quantitative metric of \emph{exposure}.
This metric
can be applied during training
as part of a testing methodology
that empirically measures
a model's potential for unintended memorization
of unique or rare sequences in the training data.

Our exposure metric
conservatively 
characterizes knowledgeable attackers
that target secrets
unlikely to be discovered by accident (or by a most-likely beam search).
As validation of this,
we describe an algorithm guided by the exposure metric
that, given a pretrained model,
can efficiently extract secret sequences
even when the model considers parts of them
to be highly unlikely.
We demonstrate our algorithm's
effectiveness in experiments,
e.g., by extracting credit card numbers
from a language model
trained on the Enron email data.
Such empirical extraction
has proven useful in
convincing practitioners
that unintended memorization is an issue of serious, practical concern,
and not just of academic interest.

Our exposure-based testing strategy is practical, 
as we demonstrate in experiments,
and by describing its use
in removing privacy risks 
for
Google's Smart Compose,
a deployed, commercial model
that is trained on millions of users' email messages
and used by other users 
for predictive text completion
during email composition~\cite{smartcompose}.

In evaluating our exposure metric,
we find unintended memorization
to be both commonplace and hard to prevent.
In particular,
such memorization is \emph{not} due to 
overtraining~\cite{OverfittingWiki}:
it occurs early during training,
and persists across different types
of models and training strategies---even
when the memorized data is very rare 
and the model size is much smaller than the
size of the training data corpus.
Furthermore,
we show that simple, intuitive 
regularization approaches such as early-stopping and dropout 
are insufficient to prevent
unintended memorization.
Only by using
differentially-private training techniques
are we able to eliminate the issue completely,
albeit at some loss in utility.

\begin{figure}
    \centering
    \vspace{-1.5em}
    \includegraphics[scale=.8]{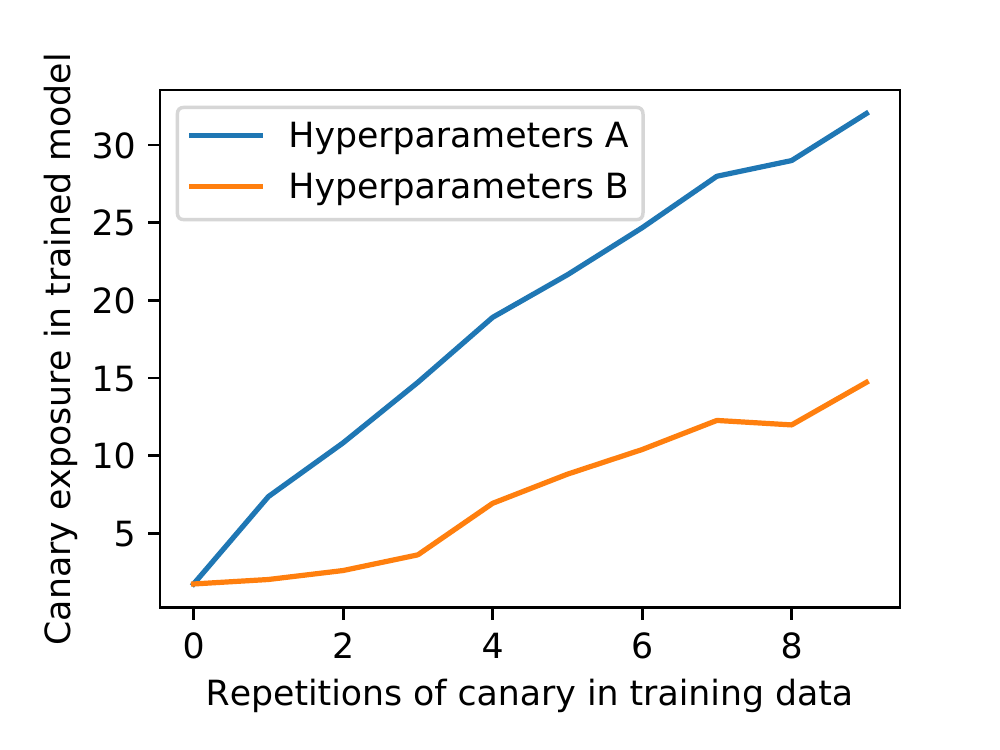}
    \caption{Results of our testing methodology applied to a state-of-the-art, word-level 
    neural-network language model~\cite{merityAnalysis}.
    Two models are trained to
    near-identical accuracy 
    using two different training strategies (hyperparameters A and B).
    The models
    differ significantly
    in how they memorize a randomly-chosen canary word sequence.
    Strategy A memorizes strongly enough
    that if the canary occurs 9 times, it can be extracted from the model
    using the techniques of Section~\ref{sec:extract}.}
    \vspace{-3ex}
    \label{fig:fig1}
\end{figure}

\paragraph{Threat Model and Testing Methodology.}
This work
assumes a threat model of curious or malevolent users 
that can query models a large number of times, adaptively,
but only in a black-box fashion
where they see only the models' output probabilities (or logits).
Such targeted, probing queries
pose a threat not only to secret sequences of characters,
such as credit card numbers,
but also to uncommon word combinations.
For example, 
if corporate data is used for training,
even simple association of words or concepts 
may reveal aspects of business strategies~\cite{AmazonPrivacy};
generative text models
can disclose even more, 
e.g., auto completing
``splay-flexed brace columns''
with the text
``using pan traps at both maiden apexes of the jimjoints,''
possibly revealing industrial trade secrets~\cite{PatriotTV}.

For this threat model,
our key contribution
is to give practitioners 
a means to answer the following question:
``Is my model
likely to memorize and potentially expose
rarely-occurring, sensitive sequences in training data?''
For this,
we describe a quantitative testing procedure
based on inserting randomly-chosen \emph{canary} sequences 
a varying number of times
into models' training data.
To gauge how much models memorize,
our exposure metric
measures the relative difference in \emph{perplexity}
between those canaries and equivalent, non-inserted random sequences.

Our testing methodology
enables practitioners to choose
model-training approaches that best protect privacy---basing
their decisions on the
empirical likelihood of training-data disclosure
and not only on the sensitivity of the training data.
Figure~\ref{fig:fig1} 
demonstrates this,
by showing
how two approaches to training a real-world model to 
the same accuracy
can dramatically differ in their unintended memorization.

\else
\input{sec1-introv2}
\fi
\section{Background: Neural Networks}
\label{sec:prelims}

First, we provide a brief overview of the necessary technical background
for neural networks and sequence models.

\subsection{Concepts, Notation, and Training}
\label{sec:nns}
A \emph{neural network} is a parameterized function $f_\theta(\cdot)$ that is designed to approximate 
an arbitrary function.
Neural networks are most often used when it is difficult to explicitly
formulate \emph{how} a function should be computed, 
but \emph{what} to compute
can be effectively specified with examples,
known as \emph{training data}.
The \emph{architecture} of the network is the general structure of the
computation, while the \emph{parameters} (or \emph{weights}) are the 
concrete internal values
$\theta$ used to compute the function.

We use standard notation \cite{goodfellow2016deep}.
Given a training set $\mathcal{X} = \{(x_i, y_i)\}_{i=1}^m$ consisting of  $m$
examples $x_i$ and labels $y_i$, the process of \emph{training} teaches
the neural network to map each given example to its corresponding
label. We train by performing (non-linear) gradient descent with
respect to the parameters $\theta$ on a
\emph{loss function} that measures how close the network is to correctly
classifying each input.
The most commonly used loss function is cross-entropy loss: given
distributions $p$ and $q$ we have
$H(p, q) = - \sum_z p(z)\log(q(z))$, 
with per-example loss
$L(x, y, \theta) = H(f_\theta(x), y)$ for $f_\theta$.

During training, we first sample a random
minibatch $\mathbb{B}$ consisting of labeled training examples $\{(\bar{x}_j,\bar{y}_j)\}_{j=1}^{m'}$
drawn from $\mathcal{X}$
(where $m'$ is the \emph{batch size}; often between 32 and 1024).
Gradient descent then updates the weights $\theta$ of the neural network by
setting
\vspace{-1em}
\[
    \theta_{\text{new}} \gets \theta_{\text{old}} - \eta {1 \over m'} \sum\limits_{j=1}^{m'} \nabla_\theta L(\bar{x}_j, \bar{y}_j, \theta)
\]
That is, we adjust the weights $\eta$-far in the direction that minimizes
the loss of the network on this batch $\mathbb{B}$ using the current weights
$\theta_{old}$. Here, $\eta$ is
called the \emph{learning rate}.

In order
to reach maximum accuracy (i.e., minimum loss),
it is often necessary to train multiple times over the entire set of training data
$\mathcal{X}$,
with each such iteration called one \emph{epoch}.
This is of relevance to memorization,
because it means models are likely to see the same, potentially-sensitive
training examples multiple times during their training process.

\subsection{Generative Sequence Models}

A generative sequence model is a fundamental architecture for
common tasks such as language-modeling \cite{bengio2003neural},
translation \cite{bahdanau2014neural}, dialogue systems, 
caption generation, optical character recognition, and automatic
speech recognition, among others.

For example, consider the task of modeling natural-language English
text from the space of all possible sequences of English words.
For this purpose, 
a generative sequence model would assign probabilities to
words based on the context in which those words appeared
in the empirical distribution of the model's training data.
For example, the model might
assign the token ``lamb'' a high
probability after seeing the sequence of words ``Mary had a little'', and the token
``the'' a low probability because---although ``the'' is a very common word---this
prefix of words requires a noun to come next,
to fit the distribution of natural, valid English.

Formally,
generative sequence models are designed to generate a sequence of tokens $x_1...x_n$ according to an (unknown) distribution
$\mathbf{Pr}(x_1...x_n)$. 
Generative sequence models estimate this distribution, which can be decomposed through Bayes' rule as
$\mathbf{Pr}(x_1...x_n)=\Pi_{i=1}^n \mathbf{Pr}(x_i|x_1...x_{i-1})$. 
Each individual computation
$\mathbf{Pr}(x_i|x_1...x_{i-1})$ represents the probability of token $x_i$ 
occurring at timestep $i$ with previous tokens $x_1$ to $x_{i-1}$.

Modern generative sequence models most frequently employ neural networks to estimate
each conditional distribution. %
To do this, a neural network is trained 
(using gradient
descent to update the
neural-network weights $\theta$)
to output the conditional probability distribution
over output tokens, given input tokens $x_1$ to $x_{i-1}$,
that maximizes the likelihood of the training-data text corpus.
For such models,
$\mathbf{Pr}(x_i|x_1...x_{i-1})$ is defined as the probability of the token $x_i$ as returned by evaluating the neural network $f_\theta(x_1...x_{i-1})$.

Neural-network generative sequence models 
most often use model architectures
that can be naturally evaluated on variable-length inputs, 
such as
Recurrent
Neural Networks (RNNs).
RNNs are evaluated using a current token (e.g., word
or character) and a current \emph{state}, and
output a predicted next token as well as an updated state.
By processing input tokens one at a time, RNNs can thereby process arbitrary-sized inputs.
In this paper we use LSTMs~\cite{hochreiter1997long} or qRNNs~\cite{bradbury2016quasi}.

\subsection{Overfitting in Machine Learning}
\label{sec:overfit}

\begin{wrapfigure}{R}{0.24\textwidth}
\vspace{-1em}
\caption{Overtraining.}
\vspace{-.6em}
\includegraphics[scale=.6,trim={.2cm 0 0 .5cm},clip]{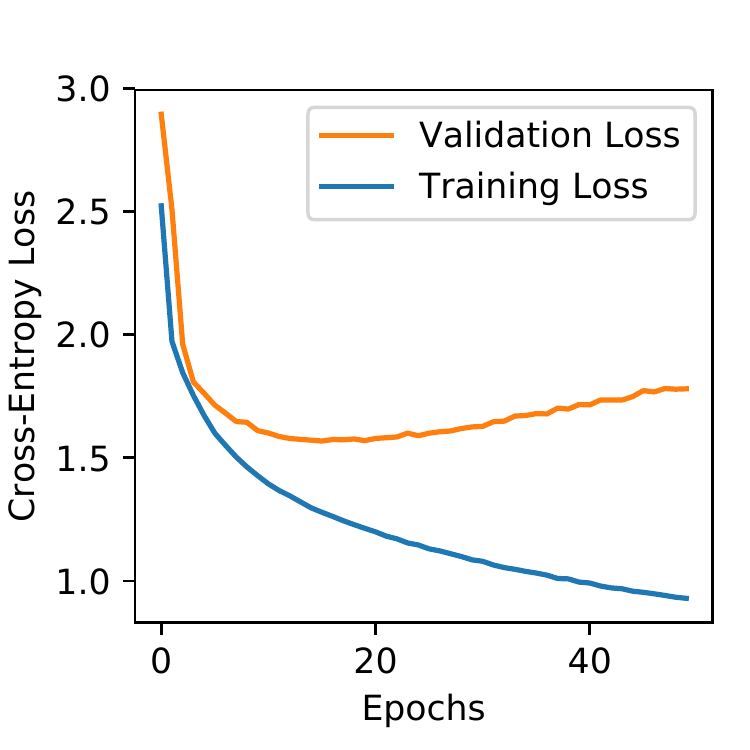}
\vspace{-2.4em}
\label{fig:overfitting}
\end{wrapfigure}
\emph{Overfitting} is one of the core difficulties in machine learning.
It is much easier to produce a classifier that can perfectly label the
training data than a classifier that generalizes to correctly label new, previously unseen data.

Because of this, when constructing a machine-learning classifier, data
is partitioned into three sets: \emph{training data}, used to train
the classifier; \emph{validation data}, used to measure the accuracy of
the classifier during training; and \emph{test data}, used only once
to evaluate the accuracy of a final classifier. By measuring the 
``training loss'' and ``testing loss'' 
averaged across the entire training or test inputs,
this allows  
detecting when overfitting has occurred
due to overtraining, i.e., training for too many steps~\cite{OverfittingWiki}.

Figure~\ref{fig:overfitting} shows a typical example of the problem of overtraining
(here
the result of training a large language model on a small dataset,
which quickly causes overfitting).
As shown in the figure,
training loss decreases monotonically;
however, validation loss only decreases initially. Once the model
has overfit the training data (at epoch 16), the validation loss begins to 
\emph{increase}.
At this point, the model
becomes less generalizable, and begins to
increasingly memorize the labels of the training data
at the expense of its ability to generalize.

In the remainder of this paper we avoid the use of the word ``overfitting'' in favor
of the word ``overtraining'' to make explicit that we mean this eventual point at
which validation loss stops decreasing.
\emph{None of our results are due to overtraining.}
Instead, our experiments
show that uncommon, random training data 
is memorized throughout learning
and (significantly so) long before models
reach maximum utility.

\section{Do Neural Nets Unintentionally Memorize?}
\label{sec:memorize}

What would it mean for
a neural network to \emph{unintentionally} memorize some of its training data? 
Machine learning must involve some form of memorization,
and even arbitrary patterns \emph{can} be memorized by neural networks 
(e.g., see~\cite{zhang2016understanding});
furthermore, the output of trained neural networks
is known to strongly suggest what training data was used
(e.g., see the membership oracle work of~\cite{shokri2017membership}).
This said, true
\emph{generalization} is the goal of
neural-network training: 
the ideal truly-general model
\emph{need not} memorize any of its training data,
especially since 
models are evaluated through their
accuracy on holdout validation data.

\paragraph{Unintended Memorization:}
The above suggests a simple definition:
unintended memorization occurs when trained neural networks
may reveal the presence of 
\emph{out-of-distribution} training data---i.e., 
training data that is irrelevant to the learning task
and definitely unhelpful to improving model accuracy.
Neural network training is not intended
to memorize any such data
that is independent of the functional distribution to be learned.
In this paper, we call such data \emph{secrets},
and our testing methodology
is based on artificially creating such secrets
(by drawing independent, random sequences from the input domain),
 inserting them as \emph{canaries} into the training data,
and evaluating their \emph{exposure} in the trained model.
When we refer to memorization without qualification, we
specifically are referring to this type of \emph{unintended} memorization.

\paragraph{Motivating Example:}
To begin, we motivate our study with a simple example that may be of practical
concern (as briefly discussed earlier).
Consider a generative sequence model trained on a text dataset used
for automated sentence completion---e.g., such one that might be used in a text-composition
assistant.
Ideally, even if the training data contained rare-but-sensitive information about some
individual users, the neural network would not memorize this information and would
never emit it as a sentence completion.
In particular, if the training data happened to contain text
written by User A with the prefix
``My social security number is ...'', one would hope that
the exact number in the suffix of User A's text 
would not be predicted
as the most-likely completion,
e.g., if User B were to type that text prefix.

Unfortunately, we show that training of neural networks can cause exactly this to occur,
unless great care is taken.
 
To make this example very concrete,
the next few paragraphs describe
the results of an experiment with a character-level
language model that predicts 
the next character
given a prior sequence of characters~\cite{mikolov2010recurrent,bengio2003neural}.
Such models are commonly used as 
the basis of everything from sentiment analysis to
compression \cite{mikolov2010recurrent,yao2017automated}. As one of the cornerstones of
language understanding, it is a representative
case study for generative modeling.
(Later, in Section~\ref{sec:nmt}, more elaborate 
variants of this experiment are described
for other types of sequence models, 
such as translation models.)

We begin by selecting a popular small dataset: the Penn Treebank (PTB) dataset 
\cite{marcus1993building}, consisting of $5$MB of text
from financial-news articles.
We train a language model on this dataset using a
two-layer LSTM with 200 hidden units (with approximately $600,\!000$  parameters).
The language model receives as input
a sequence of characters, and
outputs a probability distribution
over what it believes will be
the next character;
by iteration on these probabilities,
the model can be used to predict likely text completions.
Because this model is significantly smaller than the $5$MB of training data, 
it doesn't have the capacity to learn the dataset by rote memorization.

We augment the PTB dataset with a single out-of-distribution sentence:
``My social security number is 078-05-1120'',
and train our LSTM model on this
augmented training dataset until it reaches minimum validation loss, 
carefully doing so without any overtraining (see Section~\ref{sec:overfit}).

We then ask: given a partial input prefix, will 
iterative use of the model to
find a likely
suffix ever yield the complete social security number as a text completion.
We find the answer to our question to be an emphatic \emph{``{Yes!}''}
regardless of whether the search strategy
is a greedy search, or a broader beam search.
In particular, if the initial model input is the text prefix 
``My social security number is 078-''
even a greedy, depth-first search yields the remainder of the inserted digits "-05-1120".
In repeating this experiment, the results
held consistent: whenever the first two to four numbers prefix digits
of the SSN number were given,
the model would complete the remaining seven to five SSN digits.

Motivated by worrying results such as these,
we developed the \emph{exposure} metric, discussed next,
as well as its associated testing methodology.

\vspace{-.3em}
\section{Measuring Unintended Memorization}
\label{sec:measuring}

Having described unintentional memorization in neural networks, 
and demonstrated by empirical case study that it does sometimes occur, 
we now describe systematic methods for 
assessing the risk of disclosure due to such memorization.

\subsection{Notation and Setup}

We begin with a definition of \emph{\entropy}
that measures the likelihood of data sequences.
Intuitively, perplexity computes
the number of bits it
takes to represent 
some sequence $x$ under the distribution defined by the model~\cite{bahdanau2014neural}.

\begin{definition}{}
The \textbf{\entropy} of a sequence $x$ is
\begin{eqnarray*}
\sent_\theta(x_1...x_n)&=&-\log_2{\mathbf{Pr}(x_1...x_n | f_\theta)} \\
&=&\sum_{i=1}^n
 \bigg(-\log_2 \mathbf{Pr}(x_i | f_\theta(x_1...x_{i-1}))\bigg)
\end{eqnarray*}
\end{definition}
That is, perplexity measures
how ``surprised'' the model is to see a given value.
A higher perplexity indicates
the model is ``more surprised'' by the sequence.
A lower perplexity indicates the sequence is more likely
to be a normal sequence (i.e., perplexity is inversely correlated with
likelihood).

Naively, we might try to measure a model's unintended memorization of 
training data by directly
reporting the \entropy\ of that data.
However, whether the \entropy value is high or low 
depends heavily on the specific model, application, or dataset,
which makes the concrete
\entropy value ill suited as a direct measure of memorization. 

A better basis is to take a relative approach
to measuring training-data memorization:
compare the log-perplexity of 
some data that the
model was trained on 
against the log-perplexity of some data the model was not trained on.
While on average, models are less surprised by 
the data they are trained on,
any decent language model trained on English text
should be less surprised by (and show lower log-perplexity for) 
the phrase ``Mary had a little lamb'' than the alternate phrase
``correct horse battery staple''---even if the former never appeared in the
training data, and even if the latter \emph{did} appear in the training data.
Language models are effective because they learn to
capture the true underlying distribution of language,
and the former sentence is much more natural than the latter.
Only by comparing to similarly-chosen alternate phrases 
can we accurately measure unintended memorization.

\paragraph{Notation:}
We insert random sequences into the dataset of training data, and refer to 
those sequences as \textbf{\emph{canaries}}.\footnote{Canaries, as in ``a canary in a coal mine.''}
We create canaries based on
a \textbf{\emph{format}} sequence
that specifies how the canary sequence values are chosen randomly
using
\textbf{\emph{randomness}} $r$, from some \textbf{\emph{randomness space}}
$\mathcal{R}$.
In format sequences,
the ``holes'' denoted as $\ds{}$
are filled with random values;
for example,
the format 
$s=$ ``The random number is \ds{}\ds{}\ds{}\ds{}\ds{}\ds{}\ds{}\ds{}\ds{}''
might be filled with a
specific, random number, if  
$\mathcal{R}$ was space of digits $0$ to $9$.

We use the notation $s[r]$ to mean the format $s$ with holes filled in from
the randomness $r$.
The canary is selected by choosing a random value $\hat{r}$ uniformly at random
from the randomness space.
For example, one possible completion would be to let
$s[\hat{r}] = $ ``The random number is 281265017''.

\begin{figure}
\begin{center}
\begin{tabular}{lr}
\toprule
Highest Likelihood Sequences & Log-Perplexity \\
\midrule
\textbf{The random number is 281265017} & 14.63 \\
The random number is 281265117 & 18.56 \\
The random number is 281265011 & 19.01 \\
The random number is 286265117 & 20.65 \\
The random number is 528126501 & 20.88 \\
The random number is 281266511 & 20.99 \\
The random number is 287265017 & 20.99 \\
The random number is 281265111 & 21.16 \\
The random number is 281265010 & 21.36 \\
\end{tabular}
\captionof{table}{Possible sequences sorted by Log-Perplexity. The inserted canary--- 281265017---%
has the lowest log-perplexity. The remaining most-likely phrases are
all slightly-modified variants,
a small edit distance away from the canary phrase.\vspace*{-3ex}
}
\label{tbl:secrets}
\end{center}
\end{figure}

\subsection{The Precise Exposure Metric}
\label{sec:exposure}

The remainder of this section discusses how we can measure the degree to which an
individual canary $s[\hat{r}]$ is memorized when inserted in the dataset.
We begin with a useful definition.

\newcommand{\rank}[1]{\ensuremath{\mathbf{rank}_\theta(#1)}}

\begin{definition}{}
The \textbf{\emph{rank}} of a canary $s[r]$ is
\begin{equation*}
\rank{s[r]} = \left\lvert\{r' \in \mathcal{R} : \sent_\theta(s[r']) \le \sent_\theta(s[r]) \}\right\rvert
\end{equation*}
\end{definition}
\noindent
That is, the \emph{rank} of a specific, instantiated canary is its index in the list of all possibly-instantiated canaries,
ordered by the empirical model perplexity of all those sequences.

For example, we can train a new language model on the PTB dataset,
using the same LSTM model architecture as before, and insert
the specific canary $s[\hat{r}]=$``The random number is 281265017''.
Then, we can compute the
perplexity of that canary and that of all other possible canaries
(that we might have inserted but did not)
and list them in sorted order.
Figure~\ref{tbl:secrets} shows lowest-perplexity candidate canaries 
listed in such an experiment.\footnote{The results in this list
are not affected by the choice of the prefix text,
which might as well have been ``any random text.''
Section~\ref{sec:testing} discusses further
the impact of choosing the non-random, fixed part of the canaries' format.}
We find that the canary we insert has rank 1: no other candidate canary
$s[r']$ has lower perplexity.

The rank of an inserted canary
is \emph{not} directly linked
to the probability of generating sequences 
using greedy or beam search of
most-likely suffixes.
Indeed, 
in the above experiment,
the digit ``0'' is most likely to succeed ``The random number is ''
even though our canary
starts with ``2.''
This may prevent naive users
from accidentally finding
top-ranked sequences,
but doesn't prevent recovery
by more advanced search methods,
or even by users that know a long-enough prefix.
(Section~\ref{sec:extract} describes advanced extraction methods.)

While the rank is a conceptually useful tool for discussing the memorization of
secret data, it is computationally expensive, as it requires 
computing
the \entropy of all possible candidate canaries.
For the remainder of this section, we develop the 
concept of \textbf{exposure}: a quantity closely related to
rank, that can be efficiently approximated.

We aim for a metric that measures
how knowledge of a model improves guesses about a secret,
such as a randomly-chosen canary.
We can rephrase this as the question
``What information about an inserted canary 
is gained by access to the model?''
Thus motivated, we can
define exposure as a reduction in the
entropy of guessing canaries.

\begin{definition}{}
The \textbf{guessing entropy} is the number of guesses
$E(X)$ required in an optimal strategy to guess the value of a
discrete random variable $X$.
\end{definition}

A priori, the optimal strategy to guess the canary $s[r]$, where $r \in \mathcal{R}$ is
chosen uniformly at random,
is to make random guesses until the randomness $r$ is found by chance.
Therefore, we should expect to make $E(s[r]) = {1 \over 2}|\mathcal{R}|$
guesses before successfully guessing the value $r$.

Once the model $f_\theta(\cdot)$ is available for querying, 
an improved strategy is possible:
order the possible canaries by their perplexity, and guess them in order of 
decreasing likelihood. 
The guessing entropy for this strategy is therefore exactly
$E(s[r]\,|\,f_\theta) = \rank{s[r]}$.
Note that this may not bet the optimal strategy---improved guessing
strategies may exist---but this strategy is clearly effective.
So the reduction of work, when given access to the model $f_\theta(\cdot)$,
is given by
\[ {E(s[r]) \over E(s[r]\,|\,f_\theta)} = 
{{1\over 2}|\mathcal{R}| \over  \rank{s[r]}}. \]
Because we are often only interested in the overall scale, we
instead report the log of this value:
\begin{align*}
    \log_2 \bigg[{E(s[r]) \over E(s[r]\,|\,f_\theta)}\bigg] & =
\log_2 \bigg[{{1\over 2}|\mathcal{R}| \over  \rank{s[r]}}\bigg] \\
& = \log_2|\mathcal{R}| - \log_2 \rank{s[r]} - 1.
\end{align*}

To simplify the math in future calculations, we re-scale this value for our final
definition of exposure:

\newcommand{\exposure}[1]{\ensuremath{\mathbf{exposure}_\theta(#1)}}

\begin{definition}{}
Given a canary $s[r]$, a model with parameters $\theta$, and the randomness space $\mathcal{R}$, 
the \emph{\textbf{exposure}} of $s[r]$ is
\begin{align*}
\exposure{s[r]}  &=
\log_2{|\mathcal{R}|}-\log_2{\mathbf{rank}_\theta(s[r])}
\end{align*}

\label{def:exposure}
\end{definition}
\vspace{-1em}
Note that $|\mathcal{R}|$ is a constant. Thus the exposure is essentially computing the {\it negative log-rank} in addition to a constant to ensure the exposure is always positive.

Exposure is a real value ranging between
0 and $\log_2|\mathcal{R}|$.
Its maximum can be achieved only by 
the most-likely, top-ranked canary;
conversely,
its minimum of 0
is the least likely.
Across possibly-inserted canaries,
the median exposure is 1.

Notably, exposure is \emph{not} a normalized metric:
i.e., the magnitude of exposure values depends on the size of the search space.
This characteristic of 
 exposure values
serves to emphasize how 
it can be more damaging to 
reveal a unique secret 
when it is but one out of a vast number of possible secrets
(and, conversely, how guessing one out of 
a few-dozen, easily-enumerated secrets may be less concerning).

\subsection{Efficiently Approximating Exposure}
\label{sec:computeexposure}

We next present two approaches to approximating the exposure metric: the first a simple approach, based on sampling, and the second a more efficient, analytic approach.

\paragraph{Approximation by sampling:}
Instead of viewing exposure as measuring the reduction in
(log-scaled)
guessing entropy,
it can be viewed as measuring the excess belief that model $f_\theta$ has in
a canary $s[r]$ over random chance. 

\begin{thm} The exposure metric can also be computed as\vspace*{-1ex}
\[\exposure{s[r]}
= -\log_2 {\mathop{\mathrm{\bf Pr}}\limits_{t \in \mathcal{R}}\bigg[\big(\sent_\theta(s[t]) \le \sent_\theta(s[r])\big)\bigg]}\]\vspace*{-3ex}

\label{thm:1}
\end{thm}
\vspace*{-2ex}\emph{Proof:}\vspace*{-1ex}
\begin{align*}
\exposure{s[r]} =& \log_2{|\mathcal{R}|} - \log_2{\mathbf{rank}_\theta(s[r])} \\
=&-\log_2 {\mathbf{rank}_\theta(s[r]) \over |\mathcal{R}|} \\
=&
- \log_2{\bigg( %
     \frac{|\{t\in\mathcal{R}: \sent_\theta(s[t]) \le  \sent_\theta(s[r])\}|}{|\mathcal{R}|}}\bigg)
     \\
=& -\log_2 {\mathop{\mathrm{\bf Pr}}\limits_{t \in \mathcal{R}}\bigg[\big(\sent_\theta(s[t]) \le \sent_\theta(s[r])\big)\bigg]} 
\end{align*}

This gives us a method to approximate exposure: 
randomly choose some small
space $\mathcal{S} \subset \mathcal{R}$ (for $|\mathcal{S}| \ll |\mathcal{R}|$) 
and then compute an estimate of the exposure as
\[\exposure{s[r]} \approx - \log_2 
     {\mathop{\mathrm{\bf Pr}}\limits_{t \in \mathcal{S}}\bigg[\big(\sent_\theta(s[t]) \le \sent_\theta(s[r])\big)\bigg]}
\]

However, this sampling method is inefficient if only very few 
alternate canaries have lower entropy than $s[r]$, in which case
$|S|$ may have to be large to obtain an accurate estimate.

\paragraph{Approximation by distribution modeling:}
Using random sampling to estimate exposure
is effective when the rank of a canary is high enough (i.e. when random search
is likely to find canary candidates
$s[t]$ where 
$\sent_\theta(s[t]) \le \sent_\theta(s[r])$).
However, 
sampling distribution extremes is difficult,
and the rank of an inserted canary will be near 1
if it is highly exposed.

This is a challenging problem: given only a collection of samples, all of which have 
higher perplexity than $s[r]$, how can we estimate the number of values
with perplexity \emph{lower} than $s[r]$?
To solve it,
we can attempt to use
 \emph{extrapolation} 
 as a method to estimate exposure,
 whereas our
 previous method used \emph{interpolation}.
 
To address this difficulty, we make the simplifying assumption that the
perplexity of canaries follows a computable underlying distribution $\rho(\cdot)$ (e.g., 
a normal distribution).
To approximate $\exposure{s[r]}$, first observe
\begin{eqnarray*}
\mathop{\mathbf{Pr}}_{t \in \mathcal{R}}\big[\sent_\theta(s[t]) \le \sent_\theta(s[r])\big]\nonumber
=
\sum_{v\leq \sent_\theta(s[r])} \mathop{\mathbf{Pr}}_{t \in \mathcal{R}}\big[\sent_\theta(s[t])=v\big].
\end{eqnarray*}
Thus, from its summation form,
we can approximate the discrete distribution of \entropy using 
an integral of a continuous distribution using
\vspace{-.5em}
\begin{equation*}
\exposure{s[r]} \approx -\log_2 \int_0^{\sent_\theta(s[r])}\rho(x)\, dx
\end{equation*}
where $\rho(x)$ is a continuous density function that models the underlying
distribution of the perplexity.
This continuous distribution must allow the integral to be efficiently computed
while also
accurately approximating the distribution
$\mathbf{Pr}[\sent_\theta(s[t]) = v]$.

\begin{figure}
\vspace{-1.5em}
\centering
\includegraphics[scale=.7]{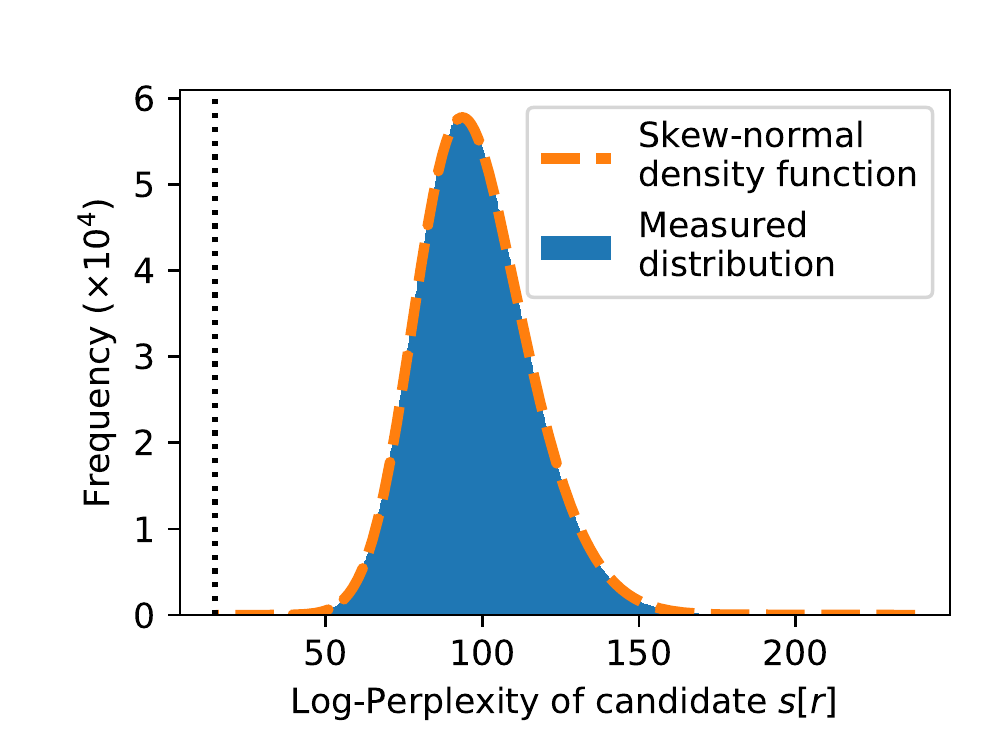}
\caption{Skew normal fit to the measured perplexity distribution.
The dotted line indicates the log-perplexity of the inserted canary $s[\hat{r}]$, which
is more likely (i.e., has lower perplexity) than any other candidate canary $s[r']$.}
\vspace{-.5em}
\label{fig:skewnormal}
\end{figure}

The above approach is an effective approximation of
the exposure metric.
Interestingly, 
this estimated exposure has no upper bound,
even though the true exposure is upper-bounded by $\log_2 |\mathcal{R}|$, when the inserted canary
is the most likely.
Usefully, this estimate can thereby
help discriminate between cases where a canary is
only marginally the most likely, and cases where
the canary is by the most likely.

In this work, we use a skew-normal distribution~\cite{o1976bayes} with mean $\mu$, standard deviation $\sigma^2$, and skew $\alpha$ to model the distribution $\rho$.
Figure~\ref{fig:skewnormal} shows a histogram 
of the \entropy of all $10^9$ different possible canaries
from our prior experiment,
overlaid with the skew-normal
distribution in dashed red. 

We observed that the
approximating skew-normal distribution almost
perfectly matches the
discrete distribution.
No statistical test can confirm that two distributions match
perfectly; instead, tests can only \emph{reject} the hypothesis
that the distributions are the same.
When we run the Kolmogorov–Smirnov goodness-of-fit test
\cite{massey1951kolmogorov} on $10^6$ samples, we fail to
reject the null hypothesis ($p>0.1$). %

\section{Exposure-Based Testing Methodology}
\label{sec:testing}

We now introduce our testing methodology which relies on the
exposure metric. The approach is simple and effective:
we have used it to discover properties about neural network memorization,
test memorization on research datasets, and test memorization of Google's
Smart Compose \cite{smartcompose}, a production model trained on billions of sequences.

The purpose of our testing methodology is to 
allow practitioners to make informed decisions based upon how much
memorization is known to occur under various settings.
For example, with this information, 
a practitioner might decide it will be necessary to apply
sound defenses (Section~\ref{sec:defense}).

Our testing strategy essentially repeats the above experiment where we
train with artificially-inserted canaries added to the training data,
and then use the exposure metric to assess to what extent the
model has memorized them.
Recall that the reason we study these fixed-format out-of-distribution
canaries is that we are focused on \emph{unintended} memorization,
and any memorization of out-of-distribution values is
by definition unintended and orthogonal to the learning task.

If, instead, we inserted in-distribution phrases which were helpful
for the learning task, then it would be perhaps even desirable
for these phrases to be memorized by the machine-learning model.
By inserting out-of-distribution phrases which we can guarantee are
unrelated to the learning task, we can measure a models propensity
to unintentionally memorize training data in a way that is not useful
for the final task.

\paragraph{Setup:} Before testing the model for memorization, we must first define
a format of the canaries that we will insert. In practice, we have found that the
exact choice of format does not significantly impact results.

However, the one choice that \emph{does} have a significant impact on the results is randomness:
it is important to choose a randomness space that matches the objective of the test to be performed.
To approximate worst-case bounds, highly out-of-distribution canaries should
be inserted; for more average-case bounds, in-distribution canaries can be used.

\paragraph{Augment the Dataset:}
Next, we instantiate each format sequence with a concrete (randomly chosen) canary by
replacing the holes $\ds{}$ with random values, e.g., words or numbers.
We then take each
canary and insert it into the training data.
In order to report detailed metrics, we can insert multiple different
canaries a varying number of times. For example, we may insert some canaries
canaries only once, some canaries tens of times, and other canaries hundreds
or thousands of times.
This allows us to establish the propensity of the model to memorize
potentially sensitive training data that may be seen a varying number of times  during training.

\paragraph{Train the Model:}
Using the same setup as will be used for training the final model, train a test model
on the augmented training data. This training process should be identical: applying
the same model using the same optimizer for the same number of iterations with the
same hyper-parameters. As we will show, each of these choices can impact the amount of
memorization, and so it is important to test on the same setup that will be used
in practice.

\paragraph{Report Exposure:}
Finally, given the trained model, we apply our exposure metric to test for
memorization. For each of the canaries, we compute and report its exposure.
Because we inserted the canaries, we will know their
format, which is needed to compute their exposure.
After training multiple models and 
inserted the same canaries a different number of times in each model, it is useful
to plot a curve showing the exposure versus the number of times that a canary has
been inserted.
Examples of such reports 
are plotted in both
Figure~\ref{fig:fig1}, shown earlier,
and Figure~\ref{fig:my_label}, shown 
on the next page.

\section{Experimental Evaluation}

This section applies our testing methodology to several model architectures and datasets
in order to (a) evaluate the efficacy of exposure as a metric, and (b) demonstrate
that unintended memorization is common across these differences.

\subsection{Smart Compose: Generative Email Model}

As our largest study, we apply our techniques to
Smart Compose \cite{smartcompose},
a generative word-level machine-learning model that is trained on a text
corpus comprising of the personal emails of millions of users.
This model has been commercially deployed for the purpose of predicting
sentence completion in email composition.
The model is in current active use by millions of users, each of which
receives predictions drawn not (only) from their own emails, but the
emails of all the users' in the training corpus.
This model is trained on highly sensitive data and its output
cannot reveal the contents of any individual user's email.

This language model
is a LSTM recurrent neural network with millions 
of parameters, trained on billions of word sequences, with a
vocabulary size of tens of thousands of words.
Because the training data contains potentially sensitive information,
we applied our exposure-based testing methodology to 
measure and ensure that only common phrases used by multiple
users were learned by the model.
By appropriately interpreting the exposure test results and limiting the
amount of information drawn from any small set of users, 
we can empirically ensure that the model is never at risk of exposing
any private word sequences from any individual user's emails.

As this is a word-level language model, our canaries
are seven (or five) randomly selected words in two formats.
In both formats the first two and
last two words are known context, and the middle %
three (or one) words vary as the randomness. 
Even with two or three words
from a vocabulary of tens of thousands,
the randomness space $\mathcal{R}$
is large enough
to support meaningful exposure measurements.

\begin{figure}
\vspace{-1em}
    \centering
    \includegraphics[scale=.7]{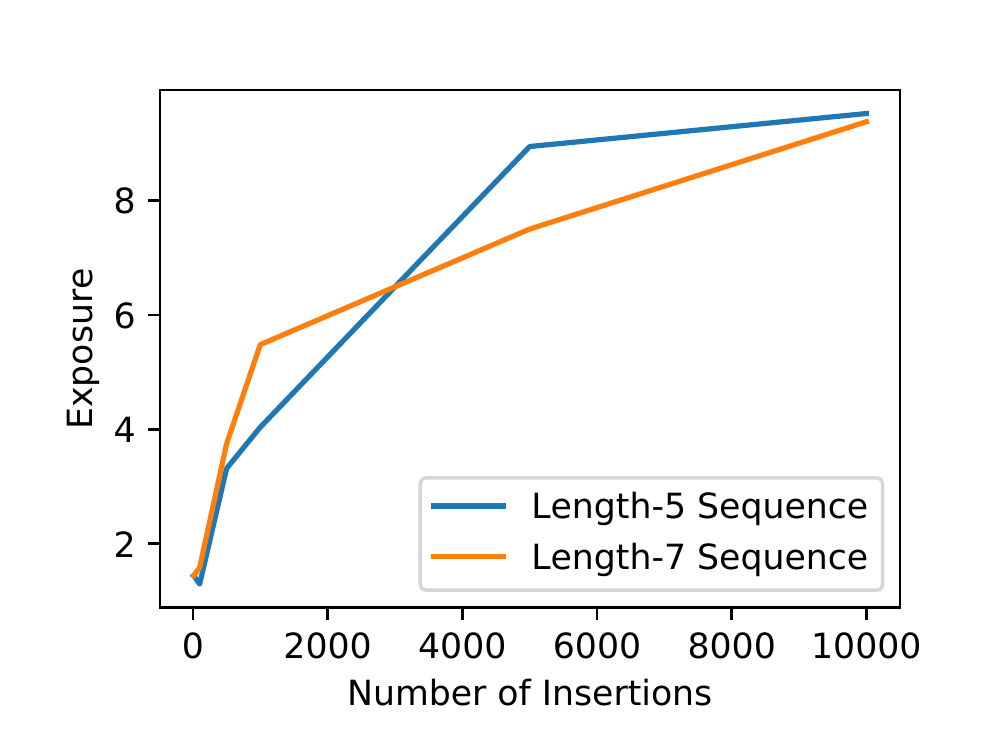}
    \caption{Exposure plot for our commercial word-level language model. Even with a canary
    inserted 10,000 times, exposure reaches only $10$: the model is
    $1,000\times$ more likely to generate this canary than another (random)
    possible phrase, but it is still not a very likely output, let alone the most likely.}
    \label{fig:my_label}
    \vspace{-1em}
\end{figure}

In more detail, we inserted multiple canaries in the training data
between 1 and 10,000 times (this does not impact model accuracy),
and trained the full
model on 32 GPUs over a billion sequences.
Figure \ref{fig:my_label} contains the results of this analysis.

(Note: The measured exposure values 
are lower than in most of 
other experiments due to the vast quantity of training data; the model is therefore exposed to the same canary
less often than in models 
trained for a large number of epochs.)

When we compute the exposure of each canary, we 
find that when secrets are very rare (i.e., one in a billion) the
model shows no signs of memorization; the measured exposure is
negligible. When the canaries are inserted
at higher frequencies, exposure
begins to increase so that the inserted canaries become
with $1000\times$ more likely than
non-inserted canaries. However,
even this higher exposure
doesn't come close to allowing discovery of canaries using our extraction algorithms (see Section~\ref{sec:extract}),
let alone accidental discovery.

Informed by these results, limits can be
placed on the incidence of unique sequences and sampling rates,
and clipping and differential-privacy noise (see Section \ref{sec:dp})
can be added to the training process, such that privacy is
empirically protected by eliminating any measured signal of exposure.

\subsection{Word-Level Language Model}
\label{sec:wordmodel}
Next we apply our technique to one of the current state-of-the-art world-level
language models \cite{merityAnalysis}.
We train this model on WikiText-103 dataset \cite{merityRegOpt}, a $500$MB cleaned subset of 
English Wikipedia.
We do not alter the open-source implementation provided by the authors; we insert
a canary five times and train the model with different hyperparameters.
We choose as a format a sequence of eight words random selected from %
the space of any of the 267,735 different words
in the model's vocabulary (i.e., that occur in the training dataset).

\begin{figure}
    \centering
    \vspace{-1.5em}
    \includegraphics[scale=.8]{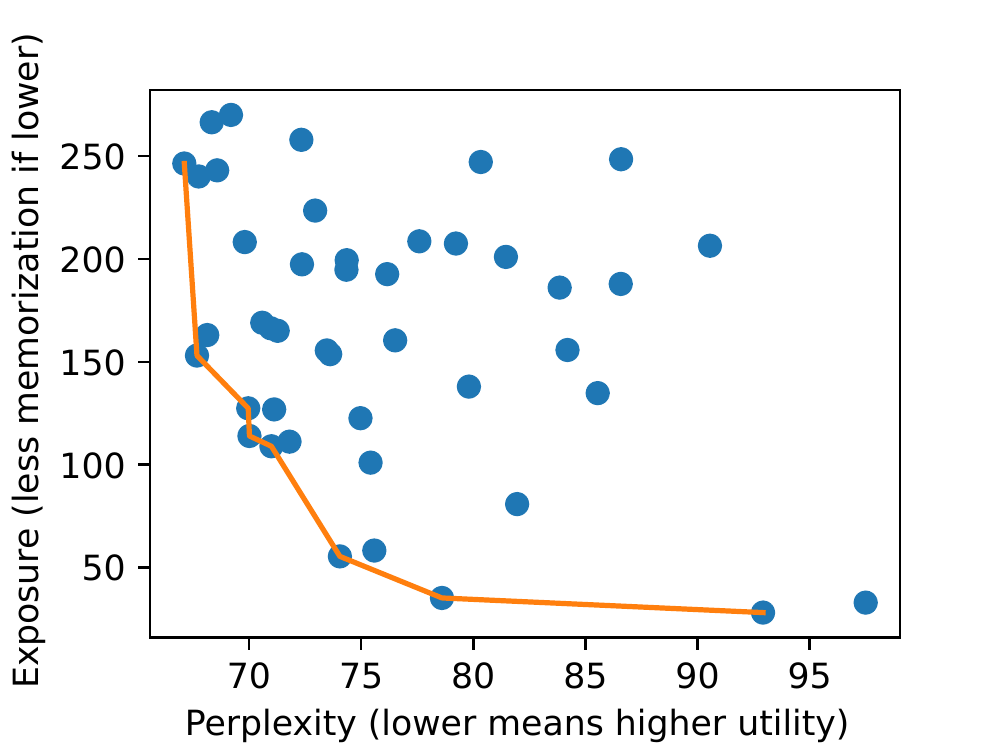}
    \caption{The results of applying our testing methodology to a word-level
    language model~\cite{merityAnalysis} inserting a
    canary five times.
    An exposure of $144$ indicates extraction should be
    possible.
    We train many models each with different hyperparameters and find
    vast differences in the level of memorization.
    The highest utility model memorizes the canary to such a degree it can be extracted.
    Other models that reach similar utility exhibit less memorization. 
    A practitioner would prefer one of the models on the Pareto frontier, which
    we highlight.}
    \vspace{-2ex}
    \label{fig:compare}
\end{figure}

We train many models with different hyperparameters and report in Figure~\ref{fig:compare}
the utility as measured by test perplexity (i.e., the exponential of the model loss)
against the measured exposure for the inserted canary.
While memorization and utility are not highly correlated (r=-0.32), this is in part due to
the fact that many choices of hyperparameters give poor utility.
We show the Pareto frontier with a solid line.

\subsection{Character-Level Language Model}
While previously we applied a small character-level model to the Penn Treebank dataset
and measured the exposure of a random number sequence, we now confirm that the results from
Section~\ref{sec:wordmodel}
hold true for a state-of-the-art character-level model.
To verify this, we apply the character-level model from \cite{merityAnalysis} to the PTB
dataset.

As expected,
based on our experiment in Section~\ref{sec:memorize}, 
we
find that a character model
model is less prone to memorizing a random
sequence of words than
a random sequence of numbers.
However, the character-level model still does memorize the inserted
random words: it reaches an exposure of $60$ (insufficient to
extract) after $16$ insertions, in contrast
to the word-models from the previous section that showed exposures much
higher than this at only $5$ insertions.

\vspace{-.5em}
\subsection{Neural Machine Translation}
\label{sec:nmt}

In addition to language modeling, another common use of 
generative sequence models is Neural Machine Translation \cite{bahdanau2014neural}.
NMT is the process
of applying a neural network to translate from one language to
another.
We demonstrate that unintentional memorization is also
a concern on this task, 
and because the domain is different, NMT also
provides us with a case study for designing a new perplexity measure.

NMT receives as input a vector of words $x_i$ in one language and outputs a vector
of words $y_i$ in a different language. It achieves this by learning an encoder
$e : \vec{x} \to \mathbb{R}^k$ that maps the input sentence to a ``thought vector''
that represents the meaning of the sentence. This $k$-dimensional vector is
then fed through a decoder $d : \mathbb{R}^k \to \vec{y}$ that decodes the thought
vector into a sentence of the target language.\footnote{See
  \cite{wu2016google} for details that we omit for brevity.}

\eat{
\begin{figure}[t]
    \centering
    \includegraphics[width=0.85\linewidth]{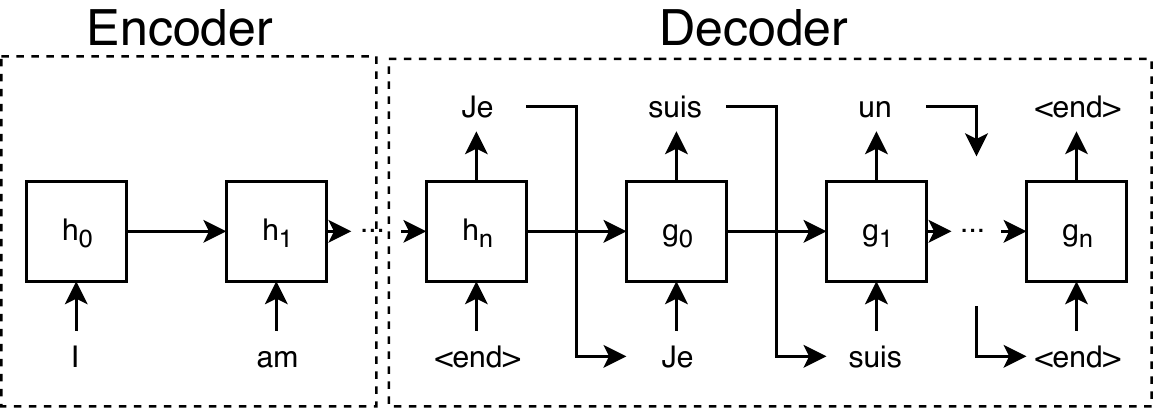}
    \caption{Diagram of the encoder-decoder architecture for neural machine translation.
    The encoder takes an input sentence and generates a ``thought vector'', which the
    decoder then translates into the target language.}
    \label{fig:nmt}
\end{figure}
}

Internally, the encoder is a recurrent neural network that maintains a
state vector and processes the input sequence one word at a time.
The final internal state is then returned as the \emph{thought vector} $v \in \mathbb{R}^k$.
 The decoder is then initialized 
with this thought vector, which the decoder uses to predict
the translated sentence one word at a time, with
every word it predicts being fed back in to generate the next.

\begin{figure}
  \vspace{-1em}
  \centering
\includegraphics[scale=.9]{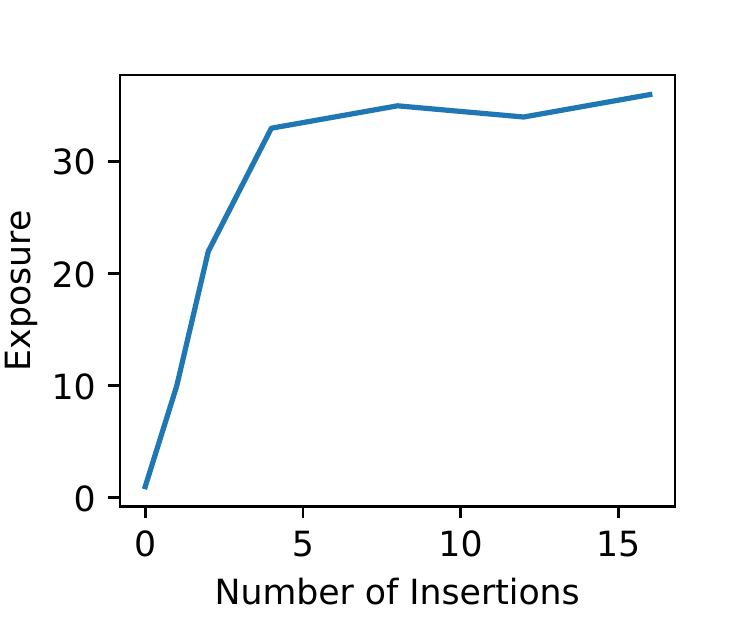}
    \caption{Exposure of a canary inserted in a Neural Machine Translation model.
      When the canary is inserted four times or more, it is
      fully memorized.}
\vspace{-1em}
\label{fig:translate}
\end{figure}

We take our NMT model directly from the TensorFlow Model Repository
\cite{tensorflow2017nmt}.
We follow the steps from the documentation to train an
English-Vietnamese model, trained on 100k sentences pairs.
We add to this dataset an English canary of the format
``My social security number is \ds{} \ds{} \ds{} - \ds{} \ds{} - \ds{} \ds{} \ds{} \ds{}''
and a corresponding Vietnamese phrase of the same format, with the English text
replaced with the Vietnamese translation, and insert this canary translation 
pair.

Because we have changed problem domains, we must define a new perplexity measure.
We feed the initial source sentence $\vec{x}$ through the encoder
to compute the thought vector.
To compute the perplexity of the source sentence mapping to the
target sentence $\vec{y}$, instead of
feeding the output of one layer to the input of the next, as we do during
standard decoding, we instead always feed $y_i$ as input to the decoder's hidden state.
The perplexity is then computed by taking the log-probability of each output
being correct, as is done on word models.
Why do we make this change to compute perplexity?
If one of the early words is guessed incorrectly and we feed it back in
to the next layer, the errors will compound and we will get an inaccurate
perplexity measure.
By always feeding in the correct output, we can accurately judge the
perplexity when changing the last few tokens.
Indeed, this perplexity definition is \emph{already implemented} in the
NMT code where it is used to evaluate test accuracy.
We re-purpose it for performing our memorization evaluation.

Under this new perplexity measure, we can now compute the exposure of the 
canary. We summarize these results in Figure~\ref{fig:translate}. By inserting
the canary only once, it already occurs $1000\times$ more
likely than random chance, and after inserting four times, it is completely
memorized.

\section{Characterizing Unintended Memorization}
\label{sec:eval:memorize}

While the prior sections clearly demonstrate
that unintended memorization \emph{is} a problem,
we now  investigate \emph{why} and \emph{how} models 
unintentionally memorize training data by
applying the testing methodology described above.

\paragraph{Experimental Setup:}
Unless otherwise specified, the experiments
in this section are performed
using the same LSTM character-level model discussed
in Section~\ref{sec:memorize} trained on the PTB dataset
with a single canary inserted with the format 
``the random number is \ds{}\ds{}\ds{}\ds{}\ds{}\ds{}\ds{}\ds{}\ds{}''
where the maximum exposure is $\log_2(10^9) \approx 30$.

\begin{figure}
\vspace{-1em}
\centering
\includegraphics[width=.95\linewidth]{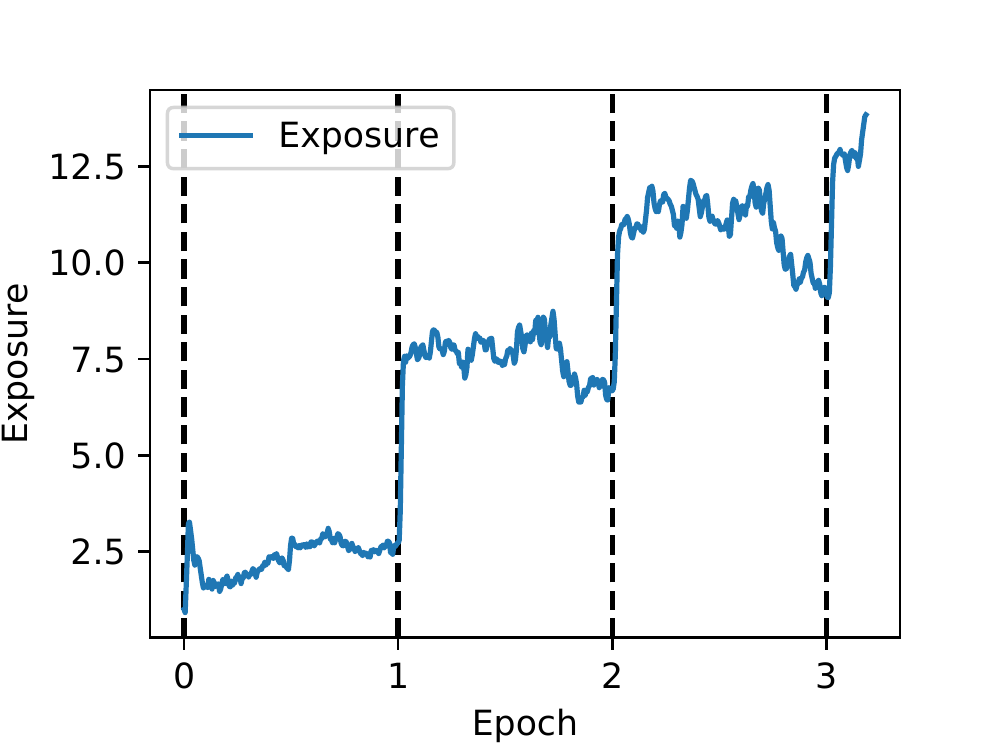}
\caption{Exposure as a function of training time. The exposure spikes after
the first mini-batch of each epoch (which contains the artificially
inserted canary),
and then falls overall during the mini-batches
that do not contain it.}
\label{fig:memovertrainshort}
\end{figure}

\subsection{Memorization Throughout Training}
To begin we apply our testing methodology to study 
a simple question: how does memorization
progress during training?

We insert the canary
near the beginning of the Penn Treebank dataset,
and disable shuffling, so that it occurs at the same point within each
epoch.
After every mini-batch
of training, we estimate the exposure of the canary. We then plot
the exposure of this canary as the training process proceeds.

Figure~\ref{fig:memovertrainshort} shows how unintended
memorization begins to occur over the
first three epochs of training on $10\%$ of the training data.
Each time the model trains on a mini-batch that contains the canary,
the exposure spikes. For the remaining mini-batches (that do not
contain the canary) the exposure randomly fluctuates and sometimes
decreases due to the randomness in stochastic gradient descent.

It is also interesting to observe that
memorization begins to occur after only
\emph{one} epoch of training: at this point, the exposure of the
canary is already 3, indicating the canary is $2^3=8\times$
more likely to occur than another random sequence chosen with the same format. 
After three epochs, the exposure is $8$: access to the model reduces
the number of guesses that would be needed to guess the canary
by over $100\times$.

\subsection{Memorization versus Overtraining}

Next, we turn to studying how unintended memorization relates to
overtraining. 
Recall we use the word \emph{overtraining} to refer to 
a form of overfitting as a result of training too long.

\begin{figure}
\vspace{-1em}
\centering
\includegraphics[width=.95\linewidth]{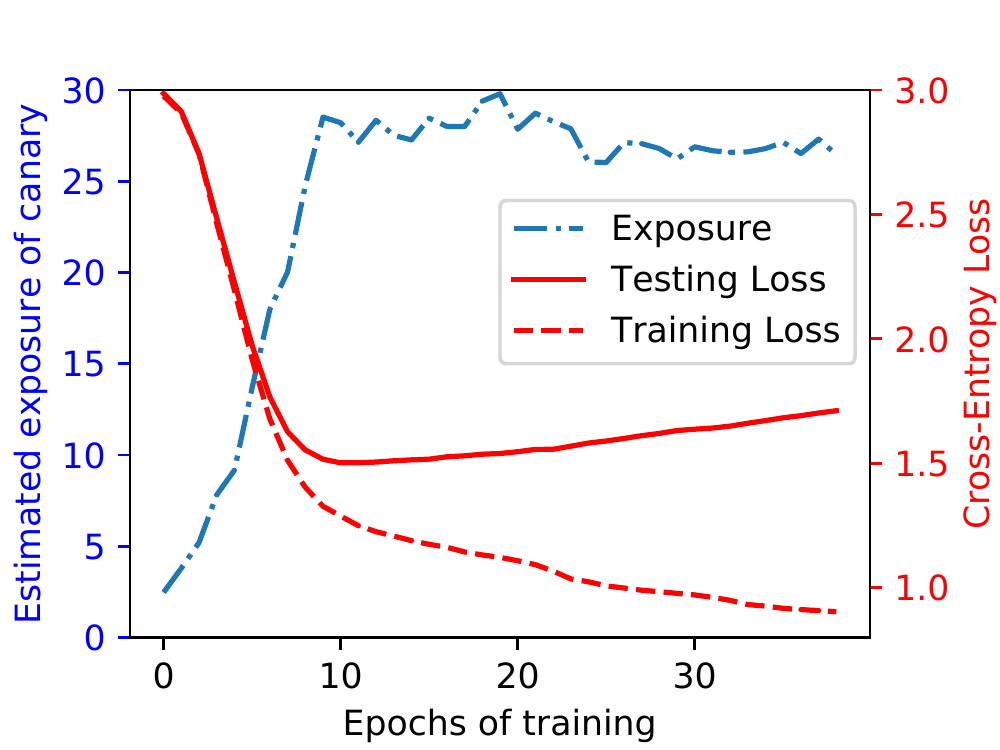}
\caption{
Comparing training and testing loss to exposure across epochs on
$5\%$ of the PTB dataset    . 
Testing loss reaches a minimum at
10 epochs, after which the model begins to over-fit (as seen by training
loss continuing to decrease).
Exposure also peaks at
this point, and decreases afterwards.}
\label{fig:memovertrain}
\end{figure}
Figure~\ref{fig:memovertrain} plots how memorization occurs during
training on a sample of $5\%$ of the PTB dataset,
so that it quickly overtrains. %
The first
few epochs see the testing loss drop rapidly,
until the minimum testing
loss is achieved at epoch 10. After this point, the testing
loss begins to increase---the model has overtrained.

Comparing this to the exposure of the canary,
we find an inverse relationship:
exposure initially increases rapidly,
until epoch 10 when the maximum amount of memorization is
achieved. Surprisingly, the exposure does not continue
increasing further, even though training continues. In fact,
the estimated exposure at epoch 10 is actually \emph{higher} than
the estimated
exposure at epoch 40 (with p-value $p<.001$).
While this is interesting, in practice it has little effect:
the rank of this canary is $1$ for all epochs after 10.

Taken together, these results are intriguing. They indicate that
unintended memorization seems to be a necessary component of training: exposure increases
when the model is learning, and does not when the model is not.
This result confirms one of the findings of Tishby and Schwartz-Ziv
\cite{shwartz2017opening} and Zhang \emph{et al.} \cite{zhang2016understanding}, who argue that neural networks first learn to
minimize the loss on the training data by memorizing it.

\subsection{Additional Memorization Experiments}

Appendix~A details some 
further memorization experiments.

\section{Validating Exposure with Extraction}
\label{sec:extract}

How accurate is the exposure metric in measuring memorization? 
We study this question by developing an \emph{extraction algorithm} that
we show can efficiently extract training data from a model when our exposure
metric indicates this should be possible (i.e., when the exposure is greater than
$\log_2{|\mathcal{R}|}$).

\subsection{Efficient Extraction Algorithm}
\paragraph{Proof of concept brute-force search:}
We begin with a simple brute-force extraction algorithm that enumerates
all possible sequences, computes their perplexity, and returns them in order
starting from the ones with lowest perplexity.
Formally, we compute $\mathop{\text{arg min}}_{r \in \mathcal{R}}\sent_\theta(s[r])$.
While this approach might be effective at validating our exposure metric accurately
captures what it means for a sequence to be memorized, 
it is unable to do so when the
space $\mathcal{R}$ is large. For example,
brute-force extraction over the space of 
credit card numbers ($10^{16}$)
would take 4,100 commodity GPU-years.

\paragraph{Shortest-path search:}
In order to more efficiently perform extraction, 
we introduce an improved search algorithm,
a modification of Dijkstra's algorithm, that
in practice reduces the complexity by several orders of magnitude.

To begin, observe it is possible to
organize all possible partial strings generated
from the \secret $s$ as a weighted tree,
where the empty string is at the root.
A partial string $b$ is a child of $a$ if $b$ expands $a$ 
by one token $t$ (which
we denote by $b = a@t$).
We set the edge weight from $a$ to $b$ to
$-\log{\mathbf{Pr}(t|f_\theta(a))}$ (i.e., the negative
log-likelihood assigned by the
model to the token $t$ following the sequence $a$).

\begin{figure}[t]
    \centering
    \includegraphics[width=0.85\linewidth]{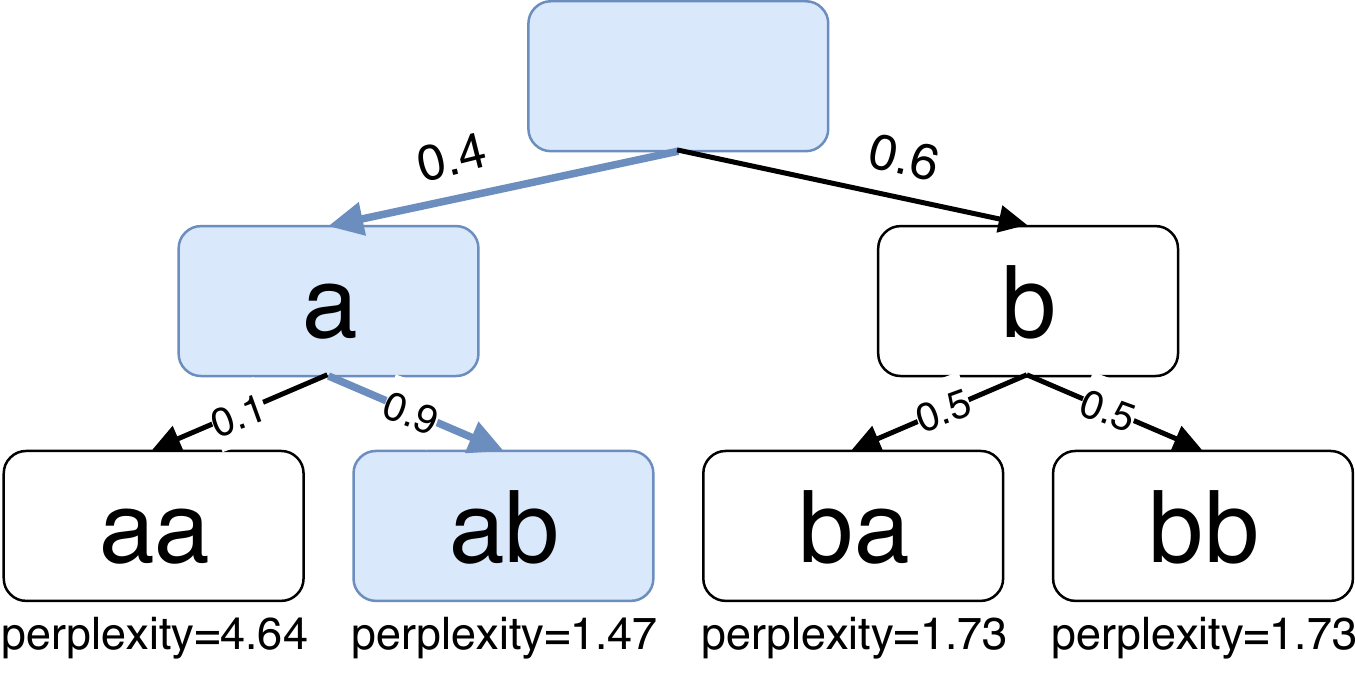}
    \caption{An example to illustrate the shortest path search algorithm. Each node
    represents one partially generated string. Each edge denotes the
     conditional probability $\mathbf{Pr}(x_i|x_1...x_{i-1})$. The path to the leaf with minimum perplexity is highlighted,
    and the log-perplexity is depicted below each leaf node.}
    \vspace{-1em}
    \label{fig:dijstra}
\end{figure}

Leaf nodes on the tree are fully-completed sequences. Observe that the
total edge weight from the root $x_1$ to a leaf node $x_n$ is given by
\begin{align*}
& \sum -\log_2{\mathbf{Pr}(x_i | f_\theta(x_1 ... x_{i-1}))} & \\
&=\sent_\theta(x_1 ... x_n) \tag{By Definition 1}
\end{align*}
Therefore, finding
$s[r]$ minimizing the cost of the path is equivalent to
minimizing its \entropy. Figure~\ref{fig:dijstra}
presents an example to illustrate the idea. Thus, finding
the sequence with lowest perplexity is equivalent to finding the lightest
path from the root to a leaf node.

Concretely, we implement a shortest-path algorithm
directly inspired by Dijkstra's algorithm~\cite{clrs}
which computes the shortest distance on a graph with
non-negative edge weights. The algorithm maintains
a priority queue of nodes on the graph. 
To initialize, only the root node (the empty string) is inserted into the priority
queue with a weight 0. In each iteration, the node with the smallest
weight is removed from the queue. Assume the node is associated with
a partially generated string $p$ and the weight is $w$. Then for
each token $t$ such that $p@t$ is a child of $p$, we insert the
node $p@t$ into the priority queue with 
$w-\log{\mathbf{Pr}(t|f_\theta(p))}$ 
where $-\log{\mathbf{Pr}(t|f_\theta(p))}$ is the weight on the edge from $p$ to $p@t$.

The algorithm terminates once the node pulled from the queue is
a leaf (i.e., has maximum length). 
In the worst-case, this algorithm may enumerate
all non-leaf nodes, (e.g., when all possible sequences have equal perplexity).
However, empirically, we find
shortest-path search enumerate from 3 to 5 orders of magnitude
fewer nodes (as we will show).

During this process, the main computational bottleneck is 
computing the edge weights $-\log{\mathbf{Pr}(t|f_\theta(p))}$.
A modern GPU can simultaneously evaluate a neural network on 
many thousand inputs in the same amount of time as it takes to evaluate one.
To leverage this benefit, we pull multiple
nodes from the priority queue at once in each iteration, and compute
all edge weights to their children simultaneously. In doing so, we observe
a $50\times$ to $500\times$ reduction in overall run-time.

Applying this optimization violates the guarantee that
the first leaf node found is always the best. We compensate
by counting the number of iterations required to find
the first full-length sequence, and continuing that many iterations more before
stopping. We then sort these sequences by \entropy and return
the lowest value. While this doubles the number of iterations, each iteration
is two orders of magnitude faster, and this results in a substantial
increase in performance.

\begin{figure}[t!]
\vspace{-2em}
\centering
\includegraphics[width=.9\linewidth]{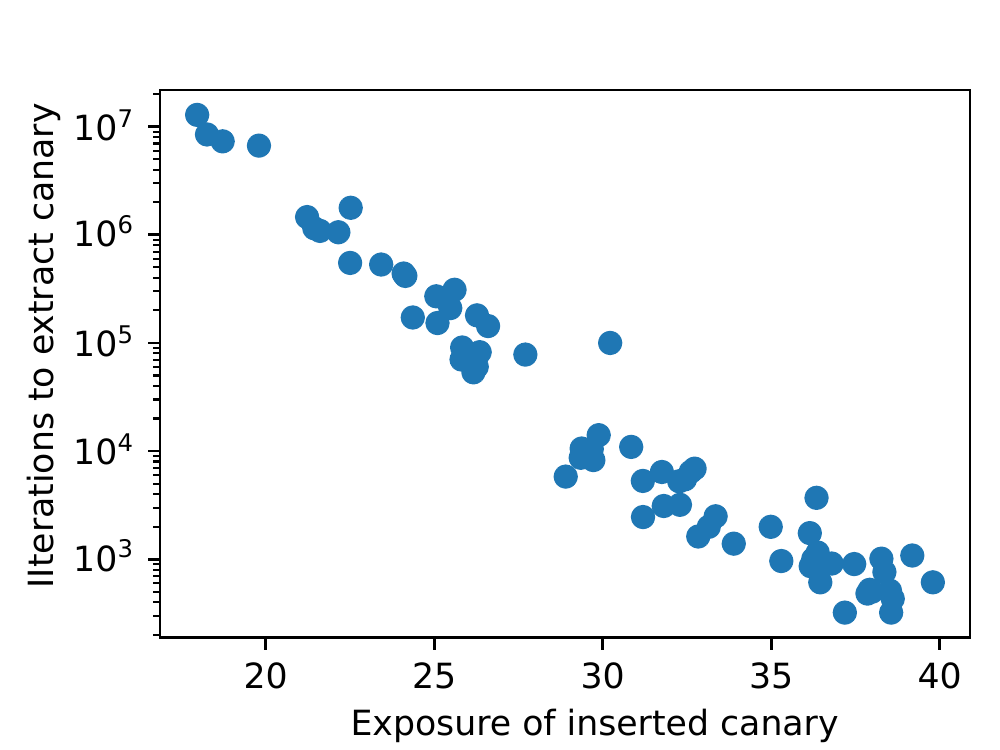}
\caption{Number of iterations the shortest-path search requires before an
inserted canary is returned, with
$|\mathcal{R}| = 2^{30}$.
At exposure 30, when the canary is fully
memorized, our algorithm requires over four orders of magnitude fewer
queries compared to brute force.}
\vspace{-.1em}
\label{fig:dijkstraiterations}
\end{figure}

\subsection{Efficiency of Shortest-Path Search}
\label{sec:eval:extract}

We begin by again using our character level language model
as a baseline, after inserting a single 9-digit random canary
to the PTB dataset once. This model completely
memorizes the canary: we find its exposure is
over 30, indicating it should be extractable.
We verify that it actually does have the lowest perplexity of all candidates
canaries by enumerating all $10^9$. %

\paragraph{Shortest path search:}
We apply our shortest-path algorithm to this model and find that it takes
only $10^5$ total queries: four orders of magnitude fewer than a brute-force
approach takes.

Perhaps as is expected, we find that the shortest-path algorithm becomes more
efficient when the exposure of the canary is higher.
We train multiple different models containing a canary to different final
exposure values (by varying model capacity and number of training epochs).
Figure~\ref{fig:dijkstraiterations} shows
the exposure of the canary versus the number
of iterations the shortest path search algorithm requires to find it.
The shortest-path search algorithm reduces the number of values enumerated
in the search from $10^9$ 
to $10^4$ (a factor of $100,000\times$ reduction) when the exposure of the inserted
phrase is greater than 30.

\subsection{High Exposure Implies Extraction}

Turning to the  main purpose of our extraction algorithm, we verify that it actually means
something when the exposure of a sequence is high.
The underlying hypothesis of our work is that exposure
is a useful measure for accurately
judging when canaries have been memorized.
We now validate that when the exposure of a phrase is high, we can extract
the phrase from the model (i.e., there are not many false positives, where exposure
is high but we can't extract it).
We train multiple models on the PTB dataset inserting a canary
(drawn from a randomness space  $|\mathcal{R}| \approx 2^{30}$)
a varying number of times with different training regimes (but train all models to the
same final test accuracy).
We then measure exposure on each of these models and attempt
to extract the inserted canary.

\begin{figure}
\vspace{-1em}
\centering
\includegraphics[width=.9\linewidth,trim=0 1ex 0 0]{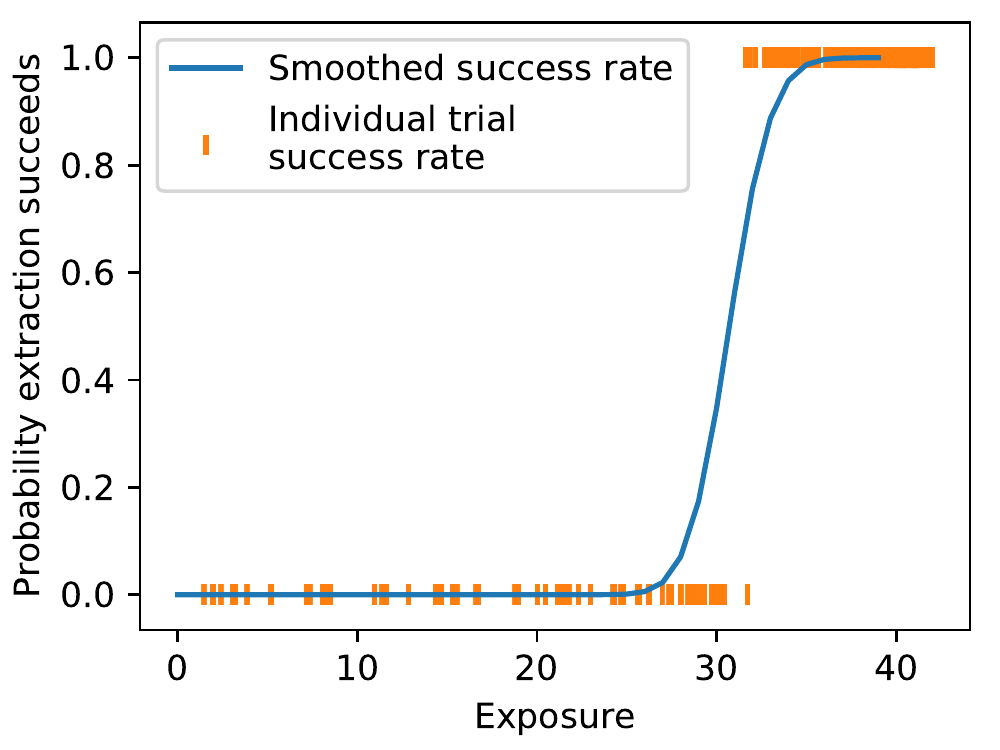}
\caption{Extraction is possible when the exposure
indicates it should be possible: when $|\mathcal{R}| = 2^{30}$, 
at an exposure of $30$ extraction quickly shifts from impossible to possible.}
\vspace{-.5em}
\label{fig:attackworks}
\end{figure}

Figure~\ref{fig:attackworks} plots
how exposure correlates with the success of extraction: extraction
is always possible when exposure is greater than $33$ but never when exposure is
less than $31$.

\subsection{Enron Emails: Memorization in Practice}
\label{sec:enron}
It is possible (although unlikely) that we detect memorization only because
we have inserted our canaries artificially. To confirm this is not the case,
we study a dataset that has many naturally-occurring secrets already in
the training data.
That is to say, instead of running experiments
on data with the canaries we have artificially inserted and treated
as ``secrets'',
we run experiments on a dataset where secrets
are pre-existing.

The Enron Email Dataset %
consists of several hundred thousand emails sent between
employees of Enron Corporation, and subsequently released by the Federal Energy
Regulatory Commission in its investigation of the company.
The complete dataset consists of the
full emails, with attachments. Many users sent highly
sensitive information in these emails, including social security numbers and
credit card numbers.

We pre-process this dataset by removing all attachments, and keep only the
body of the email. We remove the text of the email that is being responded to,
and filter out automatically-generated emails
and emails sent to the entire company.
We separate emails by sender, ranging from $1.7$MB to $5.6$MB (about the size
of the PTB dataset)
and train one character-level language model per user
who has sent at least one secret.
The language model we train is again a 2-layer LSTM, however to model the
more complex nature of writing we increase the number of units in each layer to
1024.
We again
train to minimum validation loss.%

\begin{table}[]
    \centering
    \begin{tabular}{ccrc}
         \textbf{User} & \textbf{Secret Type} & 
         \textbf{Exposure} & \textbf{Extracted?}  \\
         \midrule
         A & CCN & 52 & \checkmark \\  %
         \midrule
         B & SSN & 13 & \\ %
         \midrule
          & SSN & 16 & \\ %
         C & SSN & 10 & \\
          & SSN & 22 & \\
         \midrule
         D & SSN & 32 & \checkmark \\  %
         \midrule
         F & SSN & 13 & \\ %
         \midrule
          & CCN & 36 & \\ %
         G & CCN & 29 & \\
          & CCN & 48 &  \checkmark\\
    \end{tabular}
    \caption{Summary of results on the Enron email dataset. Three
    secrets are extractable in $<1$ hour;
    all are heavily memorized.}
    \label{tab:enronres}
\end{table}

We summarize our results in Table~\ref{tab:enronres}. Three of these
secrets (that pre-exist in the data)
are memorized to a degree that they can be extracted by
our shortest-path search algorithm.
When we run our extraction algorithm locally, 
it requires on the order of a few hours to extract
the credit card and social security numbers.
Note that it would be unfair to draw from this that an actual
attack would only take a few hours: this local attack can
batch queries to the model and does not include any remote
querying in the run-time computation.

\section{Preventing Unintended Memorization}
\label{sec:defense}

As we have shown, neural networks quickly memorize secret data.
This section evaluates (both the efficacy and impact on accuracy) three potential defenses against memorization: regularization, sanitization, and differential privacy.

\subsection{Regularization}
\label{sec:defendoverfit}

It might be reasonable to assume that unintended memorization is due to the model
{\it overtraining} to the training data. %
To show this is not the case, we apply three state-of-the-art regularization
approaches (weight 
decay~\cite{krogh1992simple}, dropout~\cite{srivastava2014dropout}, 
and quantization~\cite{hubara2016quantized})
that help prevent overtraining (and overfitting) and find that
none of these can prevent the canaries we insert from being extracted
by our algorithms.

\subsubsection{Weight Decay}
Weight decay \cite{krogh1992simple}
is a traditional approach to combat overtraining.
During training, an additional penalty
is added to the loss of the network that penalizes
model complexity.

Our initial language $600$k parameters and was trained on
the $5$MB PTB dataset. It initially does not overtrain (because it does
not have enough capacity).
Therefore, when we train our model with weight decay, we do not observe
any improvement in validation loss, or any reduction in memorization.

In order to directly measure the effect of weight decay on a model that
does overtrain, we take the first $5\%$ of the PTB dataset and train our
language model there. This time the model does overtrain the dataset without
regularization. When we add $L_2$ regularization, we see less overtraining
occur (i.e., the model reaches a \emph{lower} validation loss).
However, we observe no effect on the exposure of the canary.

\subsubsection{Dropout}
Dropout \cite{srivastava2014dropout} is a regularization
approach proposed that has been shown to
effectively prevent overtraining in neural networks. 
Again, dropout does not help with the original model on the full
dataset (and does not inhibit memorization).

We repeat the experiment above by training on $5\%$ of the data,
this time with dropout.
We vary the probability to drop a neuron from $0\%$ to $90\%$,
and train ten models at each dropout rate to eliminate
the effects of noise.

At dropout rates between $0\%$ and $20\%$, the final test accuracy of the
models are comparable (Dropout rates greater than $30\%$ reduce
test accuracy on our model). We again find that dropout does not
statistically significantly reduce the effect of unintended memorization.

\subsubsection{Quantization}
\label{app:compression}
 In our language model, each of the $600$K
parameters is represented as a 32-bit float. This puts the information
theoretic capacity of the model at $2.4$MB, which is larger than the $1.7$MB
size of the compressed PTB dataset.
To demonstrate the model is not storing a complete copy of the
training data, we show that the model can be compressed
to be much smaller while maintaining the same exposure 
and test accuracy.

To do this, we perform weight quantization \cite{hubara2016quantized}:
given a trained network with weights $\theta$, we force each weight to
be one of only $256$ different values, so each parameter can be represented
in 8 bits.
As found in prior work, quantization does not significantly affect
validation loss: our quantized model achieves a loss of $1.19$, 
compared to the original loss of $1.18$.
Additionally, we find that the exposure of the inserted canary does
not change: the inserted canary is still the most likely and is
extractable.

\subsection{Sanitization}
Sanitization is a best practice for processing sensitive, private
data.
For example, we may construct blacklists and filter out sentences containing
what may be private information from language models, or may remove all numbers
from a model trained where only text is expected.
However, one can not hope to 
guarantee that
all possible sensitive sequences
will be found and removed through such black-lists (e.g., due to the
proliferation of unknown formats or typos).

We attempted to construct an algorithm that could automatically identify potential
secrets by training two models on non-overlapping subsets of training
data and
removing any sentences where the perplexity between the two models disagreed.
Unfortunately, this style of approach missed
some secrets (and is unsound
if the same secret is inserted twice).

While sanitization is always a best practice and should be applied at every
opportunity, it is by no means a perfect defense. Black-listing is never
a complete approach in security, and so we do not consider it to be effective here.

\subsection{Differential Privacy}
\label{sec:dp}

\emph{Differential privacy}~\cite{dinur2003revealing,dwork2006calibrating,dwork2008differential} is a property
that an algorithm can satisfy which bounds the information it can leak about its inputs.
Formally defined as follows.
\begin{definition}{} A randomized algorithm $\mathcal{A}$ operating on a
dataset $\mathcal{D}$ is \emph{$(\varepsilon, \delta)$-differentially private} if
\[\mathbf{Pr}[\mathcal{A}(\mathcal{D})\in S]\leq \exp(\varepsilon)\cdot \mathbf{Pr}[\mathcal{A}(\mathcal{D}')\in S] + \delta\]
for any set $S$ of possible outputs of $\mathcal{A}$, and any two data sets  $\mathcal{D}, \mathcal{D}'$ that differ in at most one element.
\end{definition} 

Intuitively, this definition says that when adding or removing one element from
the input data set, the output distribution of a differentially private
algorithm does not change by much (i.e., by more than an
a factor exponentially small in $\varepsilon$).
Typically we set $\varepsilon=1$ and $\delta<{|\mathcal{X}|^{-1}}$ to give strong privacy
guarantees.
Thus, differential privacy is a desirable property to defend
against memorization. Consider the case where $\mathcal{D}$ contains
one occurrence of some secret training record $x$, and $\mathcal{D}'=\mathcal{D}-\{x\}$.
Imprecisely speaking, the output model of a differentially private training 
algorithm running over $\mathcal{D}$, which contains the secret, must be similar 
to the output model trained from $\mathcal{D}'$, which does not contain the secret. 
Thus, such a model can not memorize the secret as completely.

We applied the differentially-private stochastic gradient descent algorithm (DP-SGD) from \cite{abadi2016deep} 
to verify that differential privacy is an effective defense that prevents
memorization.
We used the initial, open-source code for 
DP-SGD\footnote{\scriptsize{A more modern version is at \url{https://github.com/tensorflow/privacy/}.}}
to train our character-level language model from Section~\ref{sec:memorize}.
We slightly modified this code to adapt it to recurrent neural networks 
and improved its  baseline performance by replacing the plain SGD optimizer with an RMSProp optimizer, as it often gives higher accuracy than plain SGD \cite{tieleman2012lecture}.

The DP-SGD of 
\cite{abadi2016deep} implements differential privacy by
clipping 
the per-example gradient to a max norm
and carefully adding Gaussian noise.
Intuitively, 
if the added noise matches the clipping norm,
every single, individual example
will be masked by the noise,
and cannot affect the weights of the network by itself.
As more noise is added, relative to the clipping norm, the more strict the 
$\varepsilon$ upper-bound on the privacy loss
that can be guaranteed.

\begin{table}
\centering
\begin{tabular}{llcrrrll}
& & & Test & Estimated & Extraction \\
&Optimizer & $\varepsilon$ &  Loss & Exposure & Possible? \\
\midrule
\parbox[t]{2mm}{\multirow{4}{*}{\rotatebox[origin=c]{90}{\parbox{2.5cm}{With DP}}}} \\
 & RMSProp & 0.65 & 1.69 & 1.1 &\\
 & RMSProp & 1.21 & 1.59 & 2.3 &\\
 & RMSProp & 5.26 & 1.41 & 1.8 &\\
 & RMSProp & 89 & 1.34 & 2.1 &\\
 & RMSProp & $2\times10^8$ & 1.32 & 3.2 &\\
 & RMSProp & $1\times10^9$ & 1.26 & 2.8 &\\
 & SGD & $\infty$ & 2.11 & 3.6 &\\
\parbox[t]{2mm}{\multirow{2}{*}{\rotatebox[origin=c]{90}{\parbox{1.25cm}{No DP}}}} \\
 & SGD & N/A & 1.86 & 9.5 & \\
 & RMSProp & N/A & 1.17 & 31.0 & \checkmark{} \\
 \\
 \hline
\end{tabular}
\caption{The RMSProp models trained with differential privacy do not 
memorize the training data and always have lower testing loss
than a non-private model trained using standard SGD techniques.
(Here, $\varepsilon=\infty$ indicates the moments accountant returned
an infinite upper bound on $\varepsilon$.)}
\vspace{-.5em}
\label{tab:dp-model-results}
\end{table}

We train seven differentially private models using various values of $\varepsilon$
for $100$ epochs on the PTB dataset augmented with one canary inserted.
Training a differentially
private algorithm is known to be slower than standard training; our 
implementation of this algorithm is $10-100\times$ slower than standard training.
For computing the $(\varepsilon, \delta)$ privacy budget we use the moments accountant introduced in~\cite{abadi2016deep}.
We set $\delta=10^{-9}$ in each case.
The gradient is clipped by a threshold $L=10.0$.
We initially evaluate two different optimizers (the plain SGD used by authors of~\cite{abadi2016deep} and RMSProp), but focus most experiments on training with RMSProp as we observe it achieves much better baseline results than SGD\footnote{We do
not perform hyperparameter tuning with SGD or RMSProp. SGD is known
to require extensive tuning, which may explain why it
achieves much lower accuracy (higher loss).}. 
Table~\ref{tab:dp-model-results} shows the evaluation results.

The differentially-private model with highest utility (the lowest
loss)
achieves only $10\%$ higher test loss than
the baseline model trained without differential privacy.
As we decrease $\varepsilon$ to $1.0$, the exposure drops to $1$, the point at which
this canary is no more likely than any other.
This experimentally verifies what we already expect to be true:
DP-RMSProp fully eliminates the memorization effect from a model. 
Surprisingly, however, this experiment also show that 
a little-bit of carefully-selected noise
and clipping goes a long way---as
long as the methods 
attenuate the signal from unique, secret 
input data in a principled fashion.
Even with a vanishingly-small amount of noise,
and values of $\varepsilon$ 
that offer no meaningful theoretical guarantees,
the measured exposure is negligible.

Our experience here matches that of some related work.
In particular,
other, recent measurement studies
have also found an
orders-of-magnitude
gap between
the empirical, measured privacy loss 
and the upper-bound $\varepsilon$ DP guarantees---with 
that gap growing (exponentially) as $\varepsilon$ becomes very large~\cite{EvansDP}.
Also, without modifying the training
approach, improved proof techniques have been able to 
improve guarantees by orders of magnitude,
indicating that 
the analytic $\varepsilon$ 
is not a tight upper bound.
Of course, these improved proof techniques often 
rely on additional (albeit realistic) assumptions,
such as that random shuffling can be used to provide unlinkability  \cite{erlingsson2019amplification} or that the
intermediate model weights computed during training can 
be hidden from the adversary \cite{feldman2018privacy}.
Our $\varepsilon$ calculation do not utilize these improved
analysis techniques.

\section{Related Work and Conclusions}\label{sec:conclude}
\label{sec:work}
There has been a significant amount of related work in the field of privacy and machine learning.

\paragraph{Membership Inference.}
Prior work has studied the privacy implications of training on private
data. Given
a neural network $f(\cdot)$ trained on training data $\mathcal{X}$, and an instance
$x$, it is possible to construct a \emph{membership inference attack} \cite{shokri2017membership} that answers the
question \emph{``Is $x$ a member of $\mathcal{X}$?''}.

Exposure 
can be seen as an improvement that quantifies how much memorization
has occurred (and not just \emph{if} it has).
We also show that given
only access to $f(\cdot)$, we \emph{extract} an $x$ so that
$x \in \mathcal{X}$ (and not just infer if
it is true that $x \in \mathcal{X}$), at least in the case of
generative sequence models.

Membership inference attacks have seen further study, including examining \emph{why}
membership inference is possible \cite{truex2018towards}, or mounting
inference attacks on other forms
of generative models \cite{hayes2017logan}.
Further work shows how to use membership inference attacks to
determine if a model was trained by using any individual user's personal
information \cite{song2018natural}.
These research
directions are highly important and orthogonal to ours:
this paper focuses on measuring unintended memorization, and not on any specific attacks
or membership inference queries.
Indeed, the fact that membership inference is possible is also highly
related to unintended memorization.

More closely related to our paper is 
work which produces measurements for how likely it is that membership inference
attacks will be possible \cite{long2017towards} by developing the
\emph{Differential Training Privacy} metric for cases when differentially
private training will not be possible.

\paragraph{Generalization in Neural Networks.}
Zhang \emph{et al.} \cite{zhang2016understanding} demonstrate that
standard models can be trained to perfectly fit completely random data.
Specifically, the authors show that the
same architecture that can classify MNIST
digits correctly with $99.5\%$ \emph{test
accuracy} can also be trained on completely random data to achieve $100\%$ 
\emph{train} data accuracy (but clearly poor test accuracy).
Since there is no way to learn to classify random data, 
the only explanation is that
the model has memorized all training data labels. 

Recent work has shown that overtraining can directly lead to
membership inference attacks \cite{yeom2018privacy}. %
Our work indicates that
even when we \emph{do not} overtrain our models on the training
data, unintentional memorization remains a concern.

\paragraph{Training data leakages.}
Ateniese \emph{et al.} \cite{ateniese2015hacking} show that if an adversary
is given access to a remote machine learning model
(e.g., support vector machines, hidden Markov models, neural networks, etc.)
that performs better than their own model,
it is often possible to learn information about the remote model's
training data that can be used to improve the adversary's own model. 
In this work the authors ``are not interested in privacy leaks, but
rather in discovering anything that makes classifiers better than others.''
In contrast, we focus only on the problem of private training data.

\paragraph{Backdoor (intentional) memorization.} 
Song \emph{et al.} ~\cite{song2017machine}
also study training data extraction.
The critical difference between their work and ours is that 
in their threat model, the adversary is allowed to influence
the training process and
\emph{intentionally back-doors} the model
to leak training data. They are able to achieve
incredibly powerful attacks as a result of this threat model.
In contrast, in our paper,
we show that memorization can occur, and training data leaked,
even when there is not an attacker present intentionally causing
a back-door.

\paragraph{Model stealing} studies a related problem to training data extraction:
under a black-box threat model, model stealing attempts to extract the
parameters $\theta$ (or parameters similar to them) 
from a remote model, so that the adversary can have their own copy
\cite{tramer2016stealing}.
While model extraction
is designed to steal the parameters $\theta$ of the remote model, training data
extraction is designed to extract the training data that was used to generate
$\theta$. That is, even if we were given direct access to $\theta$ 
it is still difficult to perform
training data extraction.

Later work extended model-stealing attacks to hyperparameter-stealing
attacks \cite{wang2018stealing}. These attacks are highly effective, but are
orthogonal to the problems we study in this paper.
Related work \cite{joon2018towards} also makes a similar argument that it can
be useful to steal hyperparameters in order to mount more powerful  attacks on models.

\paragraph{Model inversion} \cite{fredrikson2015model,fredrikson2014privacy} is an attack that learns aggregate statistics of the
training data, potentially revealing private information. 
For example, consider a face recognition model:
given an image of a face, it returns the probability
the input image is of some specific person.
Model inversion constructs an image that maximizes the confidence
of this classifier on the generated image; it turns out this
generated image often looks visually similar to the actual person
it is meant to classify.
No individual training instances are leaked in this
attack, only an aggregate statistic of the training data
(e.g., what the average
picture of a person looks like). In contrast, our extraction
algorithm reveals specific training examples.

\paragraph{Private Learning.}
Along with the attacks described above, there has been a large amount
of effort spent on training private machine learning algorithms. The
centerpiece of these defenses is often \emph{differential privacy}
\cite{dinur2003revealing,dwork2006calibrating,dwork2008differential,chaudhuri2009privacy,abadi2016deep}.
Our analysis in Section~\ref{sec:dp} directly follows this line of
work and we confirm that it empirically prevents the exposure
of secrets.
Other related work \cite{phong2017privacy} studies membership attacks
on differentially private training, although in the setting of a 
distributed honest-but-curious server.

Other related work \cite{nasr2018machine} studies how to apply adversarial
regularization to reduce the risk of black-box membership inference attacks,
although using different approach than taken by prior work. We do not
study this type of adversarial regularization in this paper, but believe
it would be worth future analysis in follow-up work.

\subsection{Limitations and Future Work}

This work in this paper represents a practical step towards measuring unintended
memorization in neural
networks. There are several areas where 
our work is limited in scope:
\begin{itemize}
    \item Our paper only considers generative models, as they are models that are likely
to be trained on sensitive information (credit card numbers, names, addresses, etc).
Although, our approach here will apply directly to any type of model with a defined measure of perplexity,
further work is required to handle other types of machine-learning models,
such as image classifiers.

\item Our extraction algorithm presented here was designed
to validate that canaries with a high exposure actually correspond to some
real notion of the potential to extract that canary, and by
analogy other possible secrets present in training data. However, this algorithm has assumptions
that make it ill-suited to real-world attacks.
To begin, real-world models usually only return the most likely
(i.e., the $\texttt{arg max}$) output.
Furthermore, we assume knowledge of the surrounding context 
and possible values of the canary, which may not hold true in practice.

\item Currently, we only make use of the
input-output behavior of the model to compute the exposure of sequences.
When performing our testing, we have full white-box access
including the actual weights and internal
activations of the neural network. This additional information might be used
to
develop stronger measures of memorization.
\end{itemize}
We hope future work will build on ours to develop further metrics for testing unintended memorization of unique training data details in machine-learning models.

\subsection{Conclusions}

The fact that deep learning models overfit and overtrain to their training
data has been extensively studied \cite{zhang2016understanding}.
Because neural network training
should minimize loss across all examples, training must involve a form of memorization. 
Indeed, significant machine learning research has been devoted
to developing techniques to counteract
this phenomenon \cite{srivastava2014dropout}.

In this paper we consider the related phenomenon of what we call 
\emph{unintended memorization}: deep learning models (in particular,
generative models) appear to often memorize rare details
about the training data that are completely unrelated to the
intended task while the model
is still learning the underlying behavior (i.e., while the test
loss is still decreasing).
As we show, traditional approaches to avoid overtraining
do not inhibit unintentional memorization.

Such unintended memorization of rare training details may raise
significant privacy concerns when sensitive data is used to
train deep learning models.
Most worryingly, such memorization can happen even for
examples that are present only a handful of times in the training
data, especially when those examples are outliers in the
data distribution;
this is true even for language models that make use of
state-of-the-art regularization techniques to prevent traditional
forms of overfitting and overtraining.

To date, no good method exists for helping practitioners
measure the degree to which
a model may have memorized aspects of the training data.
Towards this end,
we develop \emph{exposure}: a metric which directly
quantifies the degree to which a model has unintentionally memorized
training data.
We use exposure as the basis of a testing methodology whereby
we insert canaries (orthogonal to the learning task)
into the training data and measure their exposure.
By design, exposure is a simple metric to implement, often requiring
only a few dozen lines of code.
Indeed, our metric has, with little effort, been
applied to construct regression tests for Google's Smart Compose~\cite{smartcompose}:
a large industrial
language model trained on a privacy-sensitive text corpus.

In this way, we contribute a technique that can usefully be applied to 
aid machine learning practitioners throughout the training process, from curating
the training data, to selecting the model architecture and 
hyperparameters, all the way to extracting meaning from the
$\varepsilon$ values given by applying the
provably private techniques of differentially private
stochastic gradient descent.

\section*{Acknowledgements}
We are grateful to Martín Abadi,
Ian Goodfellow, Ilya Mironov, Ananth Raghunathan, Kunal Talwar, and David Wagner for helpful discussion
and to Gagan Bansal and the Gmail Smart Compose team
for their expertise.
We also thank our shepherd, Nikita Borisov, and  the many reviewers for their helpful suggestions.
This work was supported by National Science
Foundation award CNS-1514457, DARPA award FA8750-17-2-0091, Qualcomm, Berkeley Deep Drive,
and the Hewlett Foundation through the Center for Long-Term
Cybersecurity.
Any
opinions, findings, and conclusions or recommendations expressed
in this material are those of the author(s) and do not necessarily
reflect the views of the National Science Foundation.

\def\bibfont{\footnotesize}
{
\bibliographystyle{plain}
\bibliography{bib}
}

\appendix

\section{Additional Memorization Experiments}

\subsection{Across Different Architectures}

We evaluate different
neural network architectures in Table~\ref{tab:architecture}
again on the PTB dataset,
and find that all of them unintentionally memorize.
We observe that the two recurrent neural networks,
i.e., LSTM~\cite{hochreiter1997long} and GRU~\cite{chung2014empirical},
demonstrate both the highest accuracy (lowest loss)
and the highest exposure. 
Convolutional neural networks' accuracy
and exposure are both lower.
Therefore, through this experiment, we show that the 
memorization is not only an issue to one particular 
architecture, but appears to be common to
neural networks.

\begin{table}[t]
\centering
\begin{tabular}{lrrrr}
\toprule
Architecture & Layers & Units & Test Loss & Exposure \\
\midrule
GRU & 1 & 370 & 1.18 & 36 \\
GRU & 2 & 235 & 1.18 & 37 \\
LSTM & 1 & 320 & 1.17 & 38 \\
LSTM & 2 & 200 & 1.16 & 35 \\
CNN & 1 & 436 & 1.29 & 24 \\
CNN & 2 & 188 & 1.28 & 19 \\
CNN & 4 & 122 & 1.25 & 22 \\
WaveNet & 2 & 188 & 1.24 & 18 \\
WaveNet & 4 & 122 & 1.25 & 20 \\
\bottomrule
\end{tabular}
\caption{Exposure of a canary for various model architectures.
All models have 620K (+/- 5K) parameters and so have the same theoretical
capacity. Convolutional neural networks (CNN/WaveNet) perform less well at the language
modeling task, and memorize the canary to a lesser extent.}
\label{tab:architecture}
\end{table}

\subsection{Across Training Strategies}
There are various settings for training strategies and techniques that are known to
impact the accuracy of the final model. We briefly evaluate
the impact that each of these have on the exposure of the inserted canary.

\paragraph{Batch Size.} In stochastic gradient descent, we train
on minibatches of multiple examples simultaneously, and average their
gradients to update the model parameters. This is usually done for computational
efficiency---due to their parallel nature, modern GPUs can evaluate a neural
network on many thousands of inputs simultaneously.

To evaluate the effect of the batch size on memorization, we train our language
model with different capacity (i.e., number of LSTM units) and batch size,
ranging from $16$ to $1024$. (At each batch size for each number of units,
we train 10 models and average the results.)
All models with the same number of units reach
nearly identical final training loss and testing loss. However, the
models with larger batch size exhibit significantly more memorization,
as shown in Table~\ref{tab:batchsize}.
This experiment provides additional evidence for prior work
which has argued that using a smaller batch size yields
more generalizable models
\cite{hoffer2017train}; however we ensure that all models reach the same
final accuracy.

While this does give a method of reducing memorization for some models,
it unfortunately comes
at a significant cost: training with a small batch can be prohibitively
slow, as it may prevent parallelizing training across GPUs (and servers,
in a decentralized fashion).\footnote{%
Recent work has begun using even larger
batch sizes (e.g., $32$K) 
to train models orders of magnitude more quickly than previously 
possible~\cite{you2017scaling,goyal2017accurate}.}

\paragraph{Shuffling, Bagging, and Optimization Method.}
Given a fixed batch-size, we examine how other choices
impact memorization.
We train our model with different optimizers:
SGD, Momentum SGD,
RMSprop \cite{tieleman2012lecture}, Adagrad \cite{duchi2011adaptive}, 
Adadelta \cite{zeiler2012adadelta}, and Adam \cite{kingma2014adam};
and with either shuffling, or
bagging (where minibatches are 
sampled with replacement).

Not all models converge to the same final test accuracy. However, when we control
for the final test accuracy by taking a checkpoint from an earlier epoch from those
models that perform better, we found no statistically significant difference in
the exposure of the canary with any of these settings;
we therefore do not include these results.

\begin{table}
\centering
\begin{tabular}{rr|rrrrr}
\toprule
\multirow{9}{*}{\rotatebox[origin=c]{90}{\hspace{-3em}Batch Size}}
& & \multicolumn{5}{c}{Number of LSTM Units} \\
& & 50 & 100 & 150 & 200 & 250 \\
\midrule
& 16 & 1.7 & 4.3 & 6.9 & 9.0 & 6.4  \\
& 32 & 4.0 & 6.2 & 14.4 & 14.1 & 14.6  \\
& 64 & 4.8 & 11.7 & 19.2 & 18.9 & 21.3  \\
& 128 & 9.9 & 14.0 & 25.9 & {\bf\emph 32.5} & \textbf{\emph{35.4}}  \\
& 256 & 12.3 & 21.0 & 26.4 & {\bf{28.8}} & {\bf{31.2}}  \\
& 512 & {\bf{14.2}} & {\bf{{21.8}}} & \textbf{\emph{30.8}} & 26.0 & 26.0  \\
& 1024 & \textbf{\emph{15.7}} & \textbf{\emph{23.2}} & {\bf{{26.7}}} & 27.0 & 24.4  \\
\bottomrule
\end{tabular}
\caption{Exposure of models trained with varying model sizes
and batch sizes. Models of the same size trained for the same number of
epochs and reached similar test loss. Larger batch sizes, and larger models, both
increase the amount of memorization. The largest memorization in
each column is highlighted in italics bold, the second largest in bold.}
\label{tab:batchsize}
\end{table}

\subsection{Across Formats and Context}
We find that the context we are aware of
affects our ability to detect whether or not memorization has occurred.

In our earlier experiments we computed exposure with the
prefix ``The random number is'' and then placing the randomness as a suffix.
What if instead we knew a suffix, and the randomness was a prefix?
Alternatively, what if the randomness had a unique structure
(e.g., SSNs have dashes)?

We find that the answer is yes: additional knowledge about the format
of the canary increases our ability to detect it was memorized.
To show this, we study different insertion
formats, along with the exposure of the given canary after 5 and 10 epochs
of training in Table~\ref{tab:format}, averaged across ten models
trained with each of the formats.

\begin{table}
\centering
\begin{tabular}{l|rr}
\toprule
Format & \multicolumn{2}{c}{Exposure at Epoch} \\
& 5 & 10 \\
\midrule
\ds{}\ds{}\ds{}\ds{}\ds{}\ds{}\ds{}\ds{}\ds{} & 5.0 & 6.1  \\
$\langle$s$\rangle$ \ds{}\ds{}\ds{}\ds{}\ds{}\ds{}\ds{}\ds{}\ds{} & 6.3 & 7.1  \\
\ds{}\ds{}\ds{}\ds{}\ds{}\ds{}\ds{}\ds{}\ds{} $\langle$e$\rangle$ & 5.0 & 6.8  \\
$\langle$s$\rangle$ \ds{}\ds{}\ds{}\ds{}\ds{}\ds{}\ds{}\ds{}\ds{} $\langle$e$\rangle$ & 6.1 & 7.5 \\
\ds{}\ds{}\ds{}-\ds{}\ds{}-\ds{}\ds{}\ds{}\ds{} & 5.1 & 9.5  \\
\ds{}-\ds{}-\ds{}-\ds{}-\ds{}-\ds{}-\ds{}-\ds{}-\ds{} & 5.2 & 11.1 \\
\bottomrule
\end{tabular}
\caption{Exposure of canaries when the we are aware
of different amounts of surrounding context ($\langle$s$\rangle$ and $\langle$e$\rangle$
are in practice unique context characters of five random characters).
The exposure is computed at epoch 5 and 10, before the models
completely memorize the inserted canary.
}
\label{tab:format}
\end{table}

For the first four rows of Table~\ref{tab:format},
we use the same model, but
compute the exposure using different levels of context. This
ensures that it is only our
ability to detect exposure that changes. For the remaining two
rows, because the format has changed, we train separate
models.
We find that increasing the available context also increases the
exposure, especially when inner context is available; this
additional context becomes increasingly important as training proceeds.

\end{document}